\documentclass[10pt,twocolumn,letterpaper]{article}

\usepackage{iccv}
\usepackage{times}
\usepackage{epsfig}
\usepackage{graphicx}
\usepackage{amsmath}
\usepackage{amssymb}
\usepackage{comment}

\usepackage{soul}
\usepackage{multirow}
\usepackage[percent]{overpic}
\usepackage{xcolor}
\usepackage{anyfontsize}
\usepackage{booktabs}
\usepackage{array}
\usepackage[accsupp]{axessibility}
\newlength\Origarrayrulewidth
\newcommand{\Cline}[1]{%
 \noalign{\global\setlength\Origarrayrulewidth{\arrayrulewidth}}%
 \noalign{\global\setlength\arrayrulewidth{2pt}}\cline{#1}%
 \noalign{\global\setlength\arrayrulewidth{\Origarrayrulewidth}}%
}
\newcommand\Thickvrulel[1]{%
  \multicolumn{1}{!{\vrule width 2pt}c}{#1}%
}
\newcommand\Thickvruler[1]{%
  \multicolumn{1}{c!{\vrule width 2pt}}{#1}%
}
\usepackage{lipsum}
\newcommand\blfootnote[1]{%
  \begingroup
  \renewcommand\thefootnote{}\footnote{#1}%
  \addtocounter{footnote}{-1}%
  \endgroup
}

\usepackage[breaklinks=true,bookmarks=false]{hyperref}

\iccvfinalcopy % *** Uncomment this line for the final submission

\def\iccvPaperID{3147} % *** Enter the ICCV Paper ID here

\ificcvfinal\fi

\begin{document}

%%%%%%%%% TITLE
\title{HAA500: Human-Centric Atomic Action Dataset with Curated Videos}

\author{
\begin{tabular}{ccccc}
Jihoon Chung$^{1,2}$ & Cheng-hsin Wuu$^{1,3}$ & Hsuan-ru Yang$^1$ & Yu-Wing Tai$^{1,4}$ & Chi-Keung Tang$^1$
\end{tabular}
\\
$^1$HKUST~ $^2$Princeton University~ $^3$Carnegie Mellon University~ $^4$Kuaishou Technology
\\
{\tt\small jc5933@princeton.edu cwuu@andrew.cmu.edu hyangap@ust.hk yuwing@gmail.com cktang@cs.ust.hk}
}

\maketitle
\ificcvfinal\thispagestyle{empty}\fi

%%%%%%%%% ABSTRACT
\begin{abstract}
We contribute HAA500\footnote{
    \ificcvfinal
        HAA500 project page: {\footnotesize \url{https://www.cse.ust.hk/haa}}.
    \else
        HAA500 can be downloaded from our project website to be released.
    \fi}, a manually annotated human-centric atomic action
dataset for action recognition on 500 classes with over 591K labeled frames. To minimize ambiguities in action classification, HAA500 consists of  highly diversified classes of fine-grained atomic actions, where only consistent actions fall under the same label, e.g., ``Baseball Pitching" vs ``Free Throw in Basketball".  Thus HAA500 is different from existing
atomic action datasets, where coarse-grained atomic actions were labeled with coarse action-verbs such as ``Throw''. HAA500 has been carefully curated to capture the precise movement of human figures with little class-irrelevant motions or spatio-temporal label noises.

The advantages of HAA500 are fourfold: 1) human-centric actions with a high
average of 69.7\% detectable joints for the relevant human poses; 
2) high scalability since adding a new class 
can be done under 20--60 minutes;
3) curated videos capturing essential elements of an atomic
action without irrelevant frames;
4) fine-grained atomic action classes.
Our extensive experiments including cross-data validation using
datasets collected in the wild demonstrate the clear benefits
of human-centric and atomic characteristics of HAA500, which enable training even a baseline deep learning model to improve prediction by attending to atomic human poses.  We detail the HAA500 dataset statistics and collection methodology
and compare quantitatively with existing action recognition datasets.
\blfootnote{This work was supported by Kuaishou Technology and the
Research Grant Council of the Hong Kong SAR under grant no. 16201818.}
\end{abstract}

%%%%%%%%% BODY TEXT
\vspace{-1.5em}
\section{Introduction}

Observe the {\em coarse} annotation provided by commonly-used action recognition datasets such as~\cite{kinetics400,HMDB51,ucf101}, 
where the same action label was assigned to a given complex video action sequence (\eg, \textit{Play Soccer}, \textit{Play Baseball}) typically lasting 10 seconds or 300 frames, thus introducing a lot of ambiguities during training as two or more action categories may contain the same \textbf{atomic action}
(\eg, \textit{Run} is one of the atomic actions for both \textit{Play Soccer} and \textit{Play Baseball}).

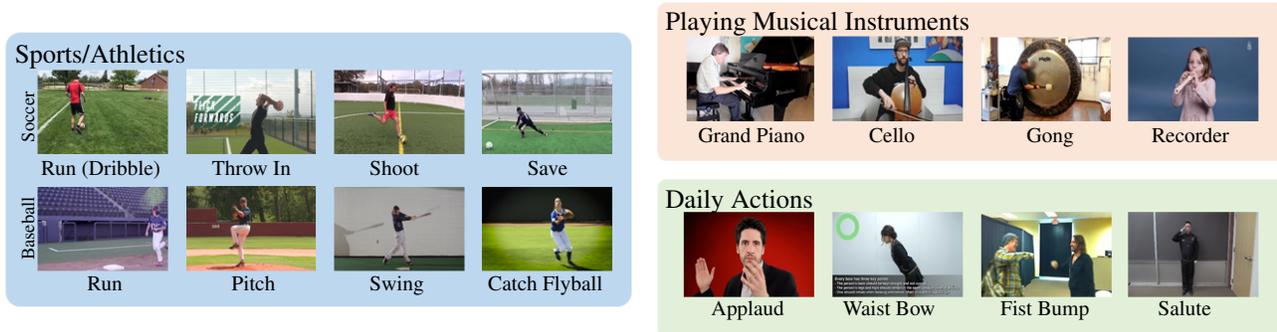
\begin{figure*}[ht!]
\begin{center}
\begin{overpic}[width=\linewidth]{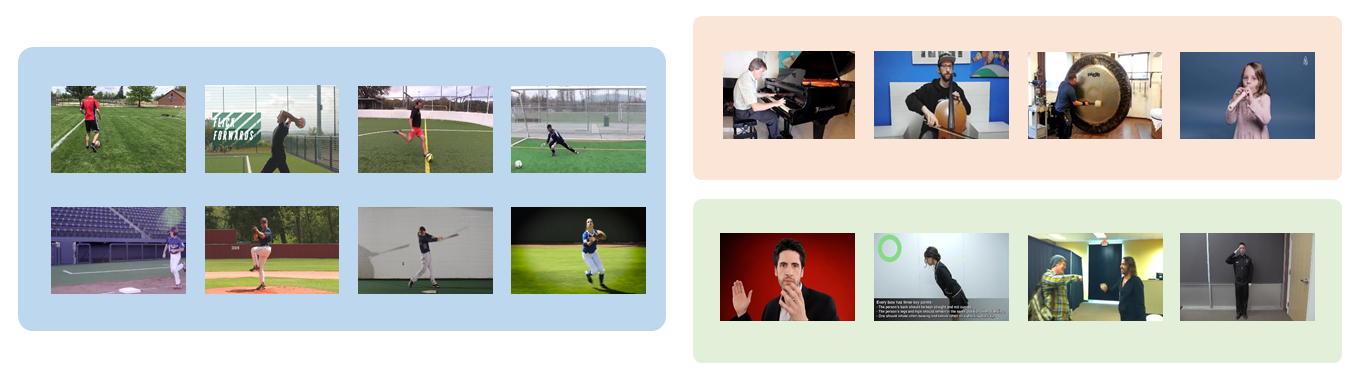}

     \put (2, 22) {Sports/Athletics}
     \put (51.5, 24.5) {Playing Musical Instruments}
     \put (51.5, 11) {Daily Actions}
     
     \put (2.5, 15.5)   {\rotatebox{90}{ \scriptsize{Soccer}}}
     \put (2.5,  6.5)   {\rotatebox{90}{ \scriptsize{Baseball}}}
     
     \put ( 4, 13.5) {\footnotesize{Run (Dribble)}}
     \put (17, 13.5) {\footnotesize{Throw In}}
     \put (29, 13.5) {\footnotesize{Shoot}}
     \put (41, 13.5) {\footnotesize{Save}}
     
     \put ( 7.5, 4.7) {\footnotesize{Run}}
     \put (18.5, 4.7) {\footnotesize{Pitch}}
     \put (29, 4.7) {\footnotesize{Swing}}
     \put (38, 4.7) {\footnotesize{Catch Flyball}}
     
     \put (54, 16) {\footnotesize{Grand Piano}}
     \put (67, 16) {\footnotesize{Cello}}
     \put (79, 16) {\footnotesize{Gong}}
     \put (88.5, 16) {\footnotesize{Recorder}}
     
     \put (55, 2.8) {\footnotesize{Applaud}}
     \put (65, 2.8) {\footnotesize{Waist Bow}}
     \put (77, 2.8) {\footnotesize{Fist Bump}}
     \put (89, 2.8) {\footnotesize{Salute}}
     
\end{overpic}
\end{center}
\vspace{-1.5em}
\caption{
    HAA500 is a fine-grained atomic action dataset, with fine-level action annotations (\eg, \textit{Soccer-Dribble}, \textit{Soccer-Throw In}) compared to the traditional composite action annotations (\eg, \textit{Soccer}, \textit{Baseball}). 
HAA500 is comparable to existing coarse-grained atomic action datasets, where we have distinctions (\eg, \textit{Soccer-Throw In}, \textit{Baseball-Pitch}) within an atomic action (\eg, \textit{Throw Something}) when the action difference is visible. The figure above displays sample videos from three different areas of HAA500. Observe that each video contains one or a few dominant human figures performing the pertinent action.
}
\vspace{-1em}
\label{fig:teaser}
\end{figure*}

Recently, atomic action datasets~\cite{ava_speech,goyal2017something,AVA,ava_speaker,finegym} have been introduced in an attempt to resolve the aforementioned issue. Google's AVA actions dataset~\cite{AVA} provides dense annotations of 80 atomic visual actions in 430 fifteen-minute video clips where actions are localized in space and time. AVA spoken activity dataset~\cite{ava_speaker} contains temporally labeled face tracks in videos, where each face instance is labeled as speaking or not, and whether the speech is audible. Something-Something dataset~\cite{goyal2017something} contains clips of humans performing pre-defined basic actions with daily objects.

However, some of their actions are still coarse which can be further split into atomic classes with significantly different motion gestures. \Eg, AVA~\cite{AVA} and Something-Something~\cite{goyal2017something} contain \textit{Play Musical Instrument} and \textit{Throw Something} as a class, respectively, where the former should be further divided into sub-classes such as \textit{Play Piano} and \textit{Play Cello}, and the latter into \textit{Soccer Throw In} and \textit{Pitch Baseball}, \etc, because each of these atomic actions has significantly different gestures. Encompassing different visual postures into a single class poses a deep neural network almost insurmountable challenge to properly learn the pertinent atomic action, which probably explains the prevailing low performance employing even the most state-of-the-art architecture, ACAR-Net (mAP: 38.30\%)~\cite{acarnet}, in AVA~\cite{AVA}, despite only having 80 classes. 

The other problem with existing action recognition video datasets is that their training examples contain actions irrelevant to the target action. 
Video datasets typically have fixed clip lengths, allowing unrelated video frames to be easily included during the data collection stage. Kinetics 400 dataset~\cite{kinetics400}, with a fixed 10-second clip length, contains a lot of irrelevant actions, \eg, showing the audience before the main \textit{violin playing}, or a person takes a long run before \textit{kicking} the ball.
Another problem is having too limited or too broad field-of-view, where a video only exhibits a part of a human interacting with an object~\cite{goyal2017something}, or a single video contains multiple human figures with different actions present~\cite{AVA,kinetics400,zhao2019hacs}.

Recently, FineGym~\cite{finegym} has been introduced to solve the aforementioned limitations by proposing fine-grained action annotations, \eg, \textit{Balance Beam-Dismount-Salto Forward Tucked}. But due to the expensive data collection process, they only contain 4 events with atomic action annotations (\textit{Balance Beam}, \textit{Floor Exercise}, \textit{Uneven Bars}, and \textit{Vault-Women}), and their clips were extracted from professional gymnasium videos in athletic or competitive events.

In this paper, we contribute Human-centric Atomic Action dataset (\textbf{HAA500}) which has been constructed with carefully curated videos with a high average of 69.7\% detectable joints, where a dominant human figure is present to perform the labeled action. The curated videos have been annotated with fine-grained labels to avoid ambiguity, and with dense per-frame action labeling and no unrelated frames being included in the collection as well as annotation.
HAA500 contains a wide variety of atomic actions, ranging from athletic atomic action~(\textit{Figure Skating - Ina Bauer}) to daily atomic action~(\textit{Eating a Burger}). 
HAA500 is also highly scalable, where adding a class takes only 20--60 minutes.
The clips are class-balanced and contain clear visual signals with little occlusion. As opposed to ``in-the-wild" atomic action datasets, our ``cultivated" clean, class-balanced dataset provides an effective alternative to advance research in atomic visual actions recognition and thus video understanding.
Our extensive cross-data experiments validate that precise annotation of fine-grained classes leads to preferable properties against datasets with orders of magnitude larger in size.

Figure~\ref{fig:teaser} shows example atomic actions collected. 

\vspace{-0.05in}

\setlength{\tabcolsep}{4pt}
\begin{table}
\begin{center}
{\small 
        \begin{tabular}{c|c|c|c}
            \hline
            Dataset & Videos &  Actions & Atomic \\ 
            \hline 
            KTH~\cite{DBLP:conf/icpr/SchuldtLC04} & 600 & 6 & \checkmark \\
            Weizmann~\cite{DBLP:conf/iccv/BlankGSIB05} & 90 & 10 & \checkmark  \\
            UCF Sports~\cite{DBLP:conf/cvpr/RodriguezAS08} & 150 & 10 &   \\
            Hollywood-2~\cite{DBLP:conf/cvpr/MarszalekLS09} & 1,707 & 12 &  \\
            HMDB51~\cite{HMDB51} & 7,000 & 51 &   \\
            UCF101~\cite{ucf101} & 13,320 & 101 &   \\
            DALY~\cite{DBLP:journals/corr/WeinzaepfelMS16} & 510 & 10 &  \\
            AVA~\cite{AVA} & 387,000 & 80 & \checkmark  \\
            Kinetics 700~\cite{kinetics700} & 650,317 & 700 & \\
            HACS~\cite{zhao2019hacs} & 1,550,000 & 200 & \checkmark  \\
            Moments in Time~\cite{momentsintime} & 1,000,000 & 339 & \checkmark\\
            FineGym~\cite{finegym} & 32,687 & 530 & \checkmark\\
            \textbf{HAA500} & \textbf{10,000} & \textbf{500} & \checkmark \\ \hline
        \end{tabular}
}
\end{center}
\caption{Summary of representative action recognition datasets.}
\vspace{-1em}
\label{table:Action_datasets}
\end{table}
% kth dataset 2004
% weizmann 2005
% ucf sports 2008
% hollywood-2 2009
% hmdb51 2011
% ucf101 2012
% daly 16
% kinetics400 2017
% hacs 2019
% mit 2019, dec
% ava 2018

\section{Related Works}

Table~\ref{table:Action_datasets} summarizes representative action recognition datasets.

\subsection{Action Recognition Dataset}
\paragraph{Composite Action Dataset}
Representative action recognition datasets, such as HMDB51~\cite{HMDB51}, UCF101~\cite{ucf101}, Hollywood-2~\cite{DBLP:conf/cvpr/MarszalekLS09}, ActivityNet~\cite{activitynet}, and Kinetics~\cite{kinetics700,kinetics400} consist of short clips which are manually trimmed to capture a single action. These datasets are ideally suited for training fully supervised, whole-clip video classifiers. A few datasets used in action recognition research, such as MSR Actions~\cite{DBLP:conf/cvpr/YuanLW09}, UCF Sports~\cite{DBLP:conf/cvpr/RodriguezAS08}, and JHMDB~\cite{DBLP:conf/iccv/JhuangGZSB13}, provide spatio-temporal annotations in each frame for short videos, but they only contain few actions. 
Aside from the subcategory of shortening the video length, recent extensions such as UCF101~\cite{ucf101}, DALY~\cite{DBLP:journals/corr/WeinzaepfelMS16}, and Hollywood2Tubes~\cite{DBLP:conf/eccv/MettesGS16} evaluate spatio-temporal localization in untrimmed videos, resulting in a performance drop due to the more difficult nature of the task. 

One common issue on these aforementioned datasets is that they are annotated with composite action classes (\eg, \textit{Playing Tennis}), thus different human action gestures (\eg, \textit{Backhand Swing}, \textit{Forehand Swing}) are annotated under a single class. Another  issue is that they tend to capture in wide field-of-view and thus include multiple human figures (\eg, tennis player, referee, audience) with different actions in a single frame, which inevitably introduce confusion to action analysis and recognition.

\vspace{-1em}
\paragraph{Atomic Action Dataset}

To model finer-level events, the AVA dataset~\cite{AVA} was introduced to provide person-centric spatio-temporal annotations on atomic actions similar to some of the earlier works~\cite{DBLP:conf/iccv/BlankGSIB05,gaidon2013temporal,DBLP:conf/icpr/SchuldtLC04}.
Other specialized datasets such as Moments in Time~\cite{momentsintime}, HACS~\cite{zhao2019hacs}, Something-Something~\cite{goyal2017something}, and Charades-Ego~\cite{sigurdsson2016hollywood} provide classes for atomic actions but none of them is a human-centric atomic action, where some of the video frames are ego-centric which only show part of a human body (\eg, hand), or no human action at all. Existing atomic action datasets~\cite{AVA,momentsintime} tend to have atomicity under English linguistics, \eg, in Moments in Time~\cite{momentsintime} \textit{Open} is annotated on video clips with a tulip opening, an eye opening, a person opening a door, or a person opening a package, which is fundamentally different actions only sharing the verb~\textit{open}, which gives the possibility of finer division.

\vspace{-1em}
\paragraph{Fine-Grained Action Dataset} 
Fine-grained action datasets try to solve ambiguous temporal annotation problems that were discussed in \cite{alwassel2018diagnosing,moltisanti2017trespassing}.
These datasets (\eg, \cite{epickitchens,jigsaws,breakfast,diving48,MPIICooking2,finegym}) use systematic action labeling to annotate fine-grained labels on a small domain of actions. Breakfast~\cite{breakfast}, MPII Cooking 2~\cite{MPIICooking2}, and EPIC-KITCHENS~\cite{epickitchens} offer fine-grained actions for cooking and preparing dishes, \eg, \textit{Twist Milk Bottle Cap}~\cite{breakfast}.
JIGSAWS~\cite{jigsaws}, Diving48~\cite{diving48}, and FineGym~\cite{finegym}  offer fine-grained action datasets respectively for surgery, diving, and gymnastics. While existing fine-grained action datasets are well suited for benchmarks, due to their low variety and the narrow domain of the classes, they cannot be extended easily in  general-purpose action recognition.

\vspace{0.1in}
Our HAA500 dataset differs from all of the aforementioned datasets as we provide a wide variety of 500 fine-grained atomic human action classes in various domains, where videos in each class only exhibit the relevant human atomic actions. 

\vspace{-0.3em}
\subsection{Action Recognition Architectures}
\vspace{-0.3em}
\setlength{\tabcolsep}{4pt}
\begin{table}[t]
\begin{center}
{\small 
        \begin{tabular}{r|c|c|c|c}
            \hline
                   & \multicolumn{2}{c|}{Kinetics 400~\cite{kinetics400}}
                   & \multicolumn{2}{c}{Something V1~\cite{goyal2017something}}\\ 
            Models & Top-1 &  Top-5 & Top-1 & Top-5  \\ 
            \hline 
            TSN (R-50)~\cite{TSN}   & 70.6\% & 89.2\% & 20.5\% & 47.5\% \\
            2-Stream I3D~\cite{i3d} & 71.6\% & 90.0\% & 41.6\% & 72.2\% \\
            TSM (R-50)~\cite{TSM}   & 74.1\% & 91.2\% & 47.3\% & 76.2\% \\
            TPN (TSM)~\cite{TPN}    & 78.9\% & 93.9\% & 50.2\% & 75.8\% \\
            \hline
            \hline Skeleton-based
                   & \multicolumn{2}{c|}{Kinetics 400~\cite{kinetics400}}
                   & \multicolumn{2}{c}{NTU-RGB+D~\cite{nturgbd}}\\ 
            Models & Top-1 &  Top-5 & X-Sub & X-View  \\ 
            \hline
            Deep LSTM~\cite{nturgbd} & 16.4\% & 35.3\% & 62.9\% & 70.3\% \\
            ST-GCN~\cite{stgcn}      & 30.7\% & 52.8\% & 81.5\% & 88.3\% \\
            \hline
        \end{tabular}
}
\end{center}
\vspace{-0.5em}
\caption{Performance of previous works on Kinetics 400~\cite{kinetics400}, Something-Something~\cite{goyal2017something}, and NTU-RGB+D~\cite{nturgbd} dataset.
We evaluate on both cross-subject (X-Sub) and cross-view (X-View) benchmarks for NTU-RGB+D.
For a fair comparison, in this paper we use~\cite{kinetics400} rather than~\cite{kinetics700} as representative action recognition model still use~\cite{kinetics400} for pre-training or benchmarking at the time of writing.   
\vspace{-1em}
\label{table:action_recognition_models}
}
\end{table}

Current action recognition architectures can be categorized into two major approaches: 2D-CNN and 3D-CNN. 2D-CNN~\cite{LRCN,feichtenhofer2016convolutional,TSM,DBLP:conf/nips/SimonyanZ14,TSN,TRN} based models utilize image-based 2D-CNN models on a single frame where features are aggregated to predict the action. While some methods (\eg, \cite{LRCN}) use RNN modules for temporal aggregation over visual features, TSN~\cite{TSN} shows that simple average pooling can be an effective method to cope with temporal aggregation. To incorporate temporal information to 2D-CNN, a two-stream structure~\cite{feichtenhofer2016convolutional,DBLP:conf/nips/SimonyanZ14} has been proposed to use RGB-frames and optical flow as separate inputs to convolutional networks.
3D-CNN~\cite{i3d,slowfast,stm} takes a more natural approach by incorporating spatio-temporal filters into the image frames. Inspired from~\cite{DBLP:conf/nips/SimonyanZ14}, two-streamed inflated 3D-CNN (I3D)~\cite{i3d} incorporates two-stream structure on 3D-CNN. SlowFast~\cite{slowfast} improves from I3D by showing that the accuracy increases when the 3D kernels are used only in the later layers of the model. A different approach is adopted in TPN~\cite{TPN} where a high-level structure is designed to adopt a temporal pyramid network which can use either 2D-CNN or 3D-CNN as a backbone. Some models~\cite{ke2017new,kim2017interpretable,stgcn} use alternative information to predict video action. Specifically, ST-GCN~\cite{stgcn} uses a graph convolutional network to predict video action from pose estimation. However, their pose-based models cannot demonstrate better performance than RGB-frame-based models.

Table~\ref{table:action_recognition_models} tabulates the performance of representative action recognition models on video action datasets, where 2D-skeleton based models~\cite{nturgbd,stgcn} show considerably low accuracy in Kinetics 400~\cite{kinetics400}.

\vspace{-0.5em}
\section{HAA500}

\subsection{Data Collection}
\vspace{-0.5em}
The annotation of HAA500 consists of two stages: vocabulary collection and video clip selection. 
While the bottom-up approach which annotates action labels on selected long videos was often used in atomic/fine-grained action datasets~\cite{AVA,finegym},
we aim to build a clean and fine-grained dataset for atomic action recognition, thus the video clips are collected based on pre-defined atomic actions following a top-down approach.

%------------------------------------------------------------------------

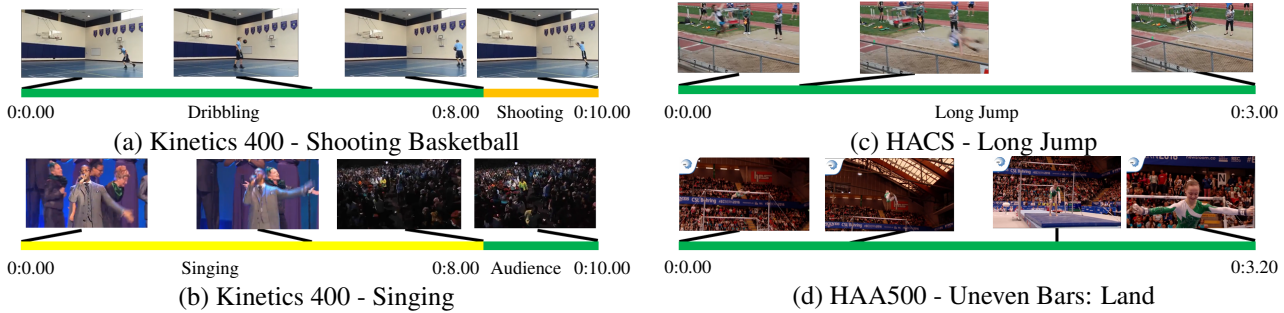
\begin{figure*}[t]
\begin{center}
	\minipage[t]{0.49\linewidth}
	    \begin{center}
	    
            \begin{overpic}[width=\linewidth]{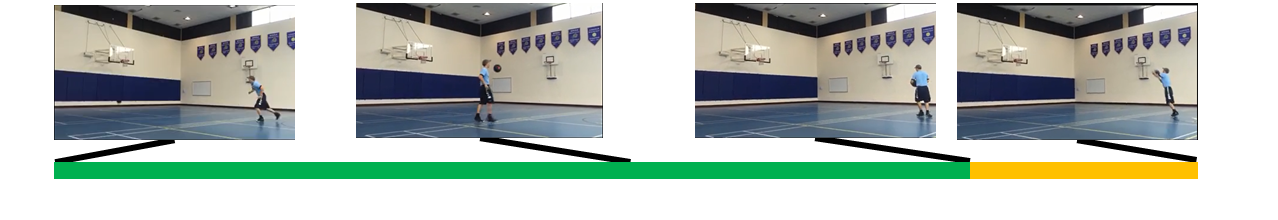}
                 \put (2,0) {\scriptsize{0:0.00}}
                 \put (30,0) {\scriptsize{Dribbling}}
                 \put (68,0) {\scriptsize{0:8.00}}
                 \put (78,0) {\scriptsize{Shooting}}
                 \put (90,0) {\scriptsize{0:10.00}}
            \end{overpic}
        	
        	(a) Kinetics 400 - Shooting Basketball
        	
            \begin{overpic}[width=\linewidth]{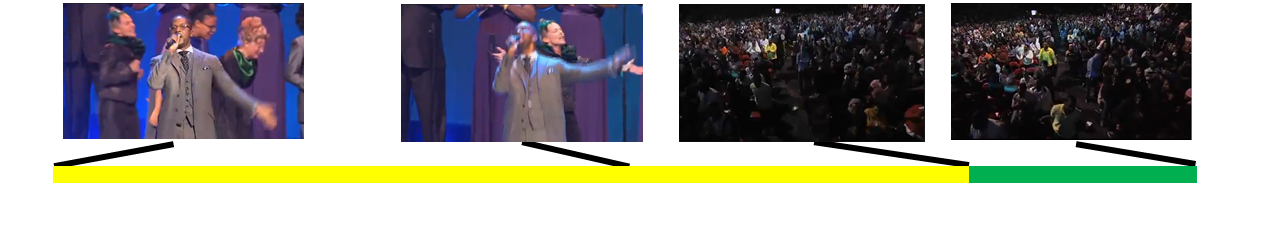}
                 \put (2,0)  {\scriptsize{0:0.00}}
                 \put (29,0) {\scriptsize{Singing}}
                 \put (68,0) {\scriptsize{0:8.00}}
                 \put (77,0) {\scriptsize{Audience}}
                 \put (90,0) {\scriptsize{0:10.00}}
            \end{overpic}
        	
        	(b) Kinetics 400 - Singing
    	\end{center}
	\endminipage\hfill
	\minipage[t]{0.49\linewidth}
	    \begin{center}
            \begin{overpic}[width=\linewidth]{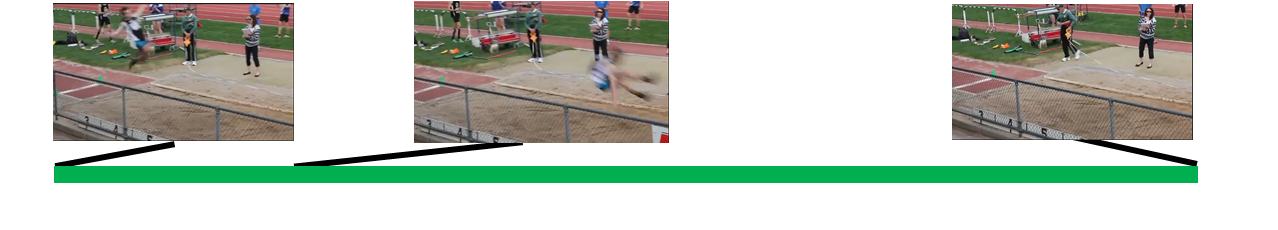}
                 \put (2,0) {\scriptsize{0:0.00}}
                 \put (44,0) {\scriptsize{Long Jump}}
                 \put (90,0) {\scriptsize{0:3.00}}
            \end{overpic}
        	\vspace{0.2em}
        	(c) HACS - Long Jump
            \begin{overpic}[width=\linewidth]{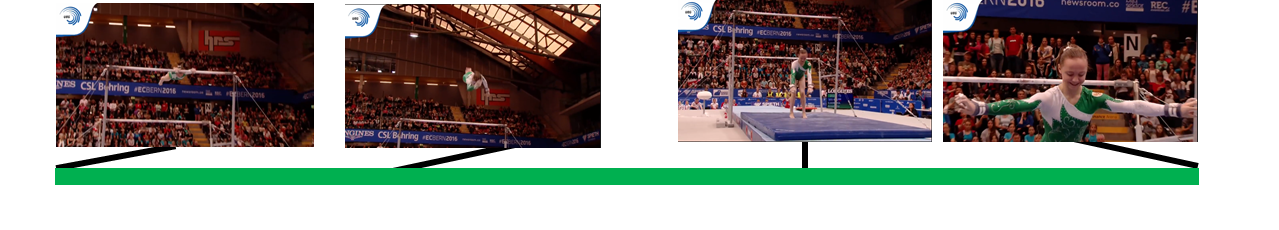}
                 \put (2,0) {\scriptsize{0:0.00}}
                 \put (90,0) {\scriptsize{0:3.20}}
            \end{overpic}
        	
        	(d) HAA500 - Uneven Bars: Land
    	\end{center}
	\endminipage\hfill
\vspace{0.5em}
\caption{Different types of label noise in action recognition datasets. 
\textbf{(a)}: Kinetics 400 has a fixed video length of 10 seconds which cannot accurately annotate quick actions like \textit{Shooting Basketball} where the irrelevant action of dribbling the ball is included in the clip. 
\textbf{(b)}: A camera cut can be seen, showing unrelated frames (audience) after the main action. 
\textbf{(c)}: By not having a frame-accurate clipping, the clip starts with a person-of-interest in the midair, and quickly disappears after few frames, causing the rest of the video clip  not to have any person in action. 
\textbf{(d)}: Our HAA500 accurately annotates the full motion of \textit{Uneven Bars - Land} without any irrelevant frames. All the videos in the class start with the exact frame an athlete puts the hand off the bar, to the exact frame when he/she finishes the landing pose. }
\label{fig:comparison_noise}
\end{center}
\vspace{-2em}
\end{figure*}

\vspace{-1em}
\subsubsection{Vocabulary Collection}
\vspace{-0.5em}
To make the dataset as clean as possible and useful for recognizing fine-grained atomic actions, we narrowed down the scope of our super-classes into 4 areas; \textit{Sport/Athletics}, \textit{Playing Musical Instruments}, \textit{Games and Hobbies}, and \textit{Daily Actions}, where future extension beyond the existing classes is feasible. 
We select action labels where the variations within a class are typically indistinguishable. For example, instead of \textit{Hand Whistling}, we have \textit{Whistling with One Hand} and \textit{Whistling with Two Hands}, as the variation is large and distinguishable. 
Our vocabulary collection methodology makes the dataset hierarchical where atomic actions may be combined to form a composite action, \eg, \textit{Whistling} or \textit{Playing Soccer}. 
Consequently, HAA500 contains 500 atomic action classes, where 212 are \textit{Sport/Athletics}, 51 are \textit{Playing Musical Instruments}, 82 are \textit{Games and Hobbies} and 155 are \textit{Daily Actions}.

%------------------------------------------------------------------------
% Dataset Summary

\begin{table}[t]
{\small
\setlength\extrarowheight{3pt}
\begin{center}
{\small 
\begin{tabular}{|c c c c c|}
\Cline{1-5}
\Thickvrulel{action} & clips & mean length & duration & \Thickvruler{frames} \\ \Cline{1-5}
500 & 10,000 & 2.12s & 21,207s & 591K \\ \hline
\addlinespace
\Cline{0-3}
\Thickvrulel{no. of people} & 1 & 2 & \Thickvruler{$>$2} \\ \Cline{0-3}
             & 8,309 & 859 & \multicolumn{1}{c|}{832} \\ \cline{0-3}
\addlinespace
\Cline{0-2}
\Thickvrulel{moving camera} & O & \Thickvruler{X} \\ \Cline{0-2}
              & 2,373 & \multicolumn{1}{c|}{7,627} \\ \cline{0-2}
\end{tabular}
}
\end{center}
}
\caption{Summary of HAA500.}
\label{table:dataset_summary}
\vspace{-0.2in}
\end{table}

\vspace{0em}
\subsubsection{Video Clip Selection}
To ensure our dataset is clean and class-balanced, all the video clips are collected from YouTube with the majority having a resolution of at least 720p and each class of atomic action containing 16 training clips.
We manually select the clips with apparent human-centric actions where the person-of-interest is the only dominant person in the frame at the center with their body clearly visible. 
To increase diversity among the video clips and avoid unwanted bias, all the clips were collected from different YouTube videos, with different environment settings so that the action recognition task cannot be trivially reduced to identifying the corresponding backgrounds. Clips are properly trimmed in a frame-accurate manner to cover the desired actions while assuring every video clip to have compatible actions within each class (\eg, every video in the class \textit{Salute} starts on the exact frame where the person is standing still before moving the arm, and the video ends when the hand goes next to the eyebrow). Refer to Figure~\ref{fig:teaser} again for examples of the selected videos.

\vspace{-1em}
\subsubsection{Statistics}
Table~\ref{table:dataset_summary} summarizes the HAA500 statistics. HAA500 includes 500 atomic action classes where each class contains 20 clips, with an average length of 2.12 seconds. 
Each clip was annotated with meta-information which contains the following two fields: the number of dominant people in the video and the camera movement.

\vspace{-1em}
\subsubsection{Training/Validation/Test Sets}
Since the clips in different classes are mutually exclusive, all clips appear only in one split. The 10,000 clips are split as 16:1:3, resulting in segments of 8,000 training, 500 validation, and 1,500 test clips.

\begin{figure*}[t]
\begin{center}
\begin{overpic}[width=\linewidth]{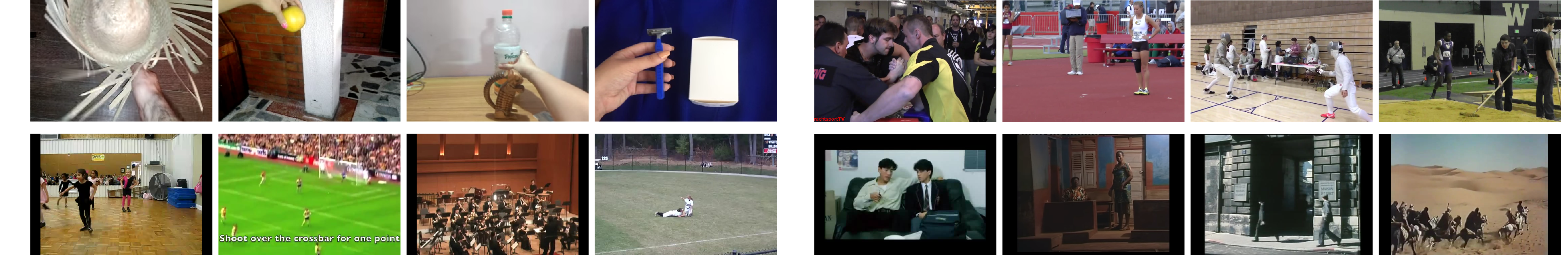}
 \put (0.4,-0.1)   {\rotatebox{90}{ {\fontsize{6}{6}\selectfont Kinetics 400}}}
 \put (50.4,2.1)  {\rotatebox{90}{ {\fontsize{6}{6}\selectfont AVA}}}
 \put (0.4, 9)  {\rotatebox{90}{ {\fontsize{6}{6}\selectfont Something}}}
 \put (50.4, 10.4) {\rotatebox{90}{ {\fontsize{6}{6}\selectfont HACS}}}
\end{overpic}
\caption{The video clips in AVA, HACS, and Kinetics 400 contain multiple human figures with different actions in the same frame. Something-Something focuses on the target object and barely shows any human body parts. In contrast, all video clips in HAA500
are carefully curated where each video shows either a single person or the person-of-interest as the most dominant figure in a given frame.}
\label{fig:comparison_human_centric}
\end{center}
\vspace{-0.3in}
\end{figure*}
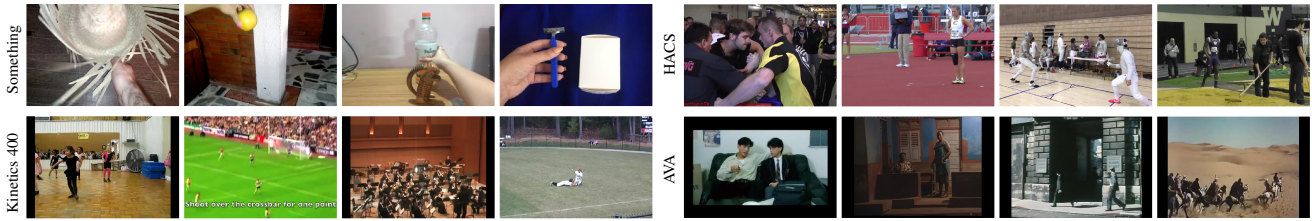   

\setlength{\tabcolsep}{4pt}
\begin{table}[t]
\begin{center}
{\small 
\begin{tabular}{c|c|c|c}
\hline 
Dataset                     & Clip Length & Irr. Actions & Camera Cuts    \\ \hline
UCF101~\cite{ucf101}        & Varies      &                 &             \\ 
HMDB51~\cite{HMDB51}        & Varies      &                 & \checkmark  \\ 
AVA~\cite{AVA}              & 1 second    & \checkmark      & \checkmark  \\ 
HACS~\cite{zhao2019hacs}    & 2 second    & \checkmark      &             \\ 
Kinetics~\cite{kinetics400}    & 10 second   & \checkmark      & \checkmark  \\ 
M.i.T.~\cite{momentsintime} & 3 second    &                 &             \\ 
\textbf{HAA500}             & Just Right  &                 &             \\ \hline
\end{tabular}
}
\end{center}
\caption{Clip length and irrelevant frames of video action datasets.}
\vspace{-1em}
\label{table:comparison_sampling_rate}

\end{table}
\setlength{\tabcolsep}{1.4pt}

\subsection{Properties and Comparison}

\subsubsection{Clean Labels for Every Frame}

Most video datasets~\cite{AVA,kinetics400,ucf101} show strong label noises, due to the difficulties of collecting clean video action datasets. Some~\cite{kinetics400,HMDB51,ucf101} often focus on the ``scene" of the video clip, neglecting the human ``action" thus including irrelevant actions or frames with visible camera cuts in the clip. Also, video action datasets~\cite{AVA,kinetics400,momentsintime,zhao2019hacs} have fixed-length video clips, so irrelevant frames are inevitable for shorter actions. Our properly trimmed video collection guarantees a clean label for every frame. 

Table~\ref{table:comparison_sampling_rate} tabulates clip lengths and label noises of video action datasets. 
Figure~\ref{fig:comparison_noise} shows examples of label noises. As HAA500 is constructed with accurate temporal annotation in mind, we are almost free from any adverse effects due to these noises.

\vspace{-0.5em}
\subsubsection{Human-Centric}

One potential problem in action recognition is that the neural network may predict by trivially comparing the background scene in the video, or detecting key elements in a frame (\eg, a basketball to detect \textit{Playing Basketball}) rather than recognizing the pertinent human gesture, thus causing the action recognition to have no better performance improvements over scene/object recognition. The other problem stems from the video action datasets where videos captured in wide field-of-view contain multiple people in a single frame~\cite{AVA,kinetics400,zhao2019hacs}, while videos captured using narrow field-of-view only exhibit very little body part in interaction with the pertinent object~\cite{goyal2017something,momentsintime}. 

In~\cite{AVA} attempts were made to overcome this issue through spatial annotation of each individual in a given frame. This introduces another problem of action localization and thus further complicating the difficult recognition task. Figure~\ref{fig:comparison_human_centric} illustrates example frames of different video action datasets. 

HAA500 contributes a curated dataset where human joints can be clearly detected over any given frame, thus allowing the model to benefit from learning human movements than just performing scene recognition. As tabulated in Table~\ref{table:comparison_joint}, HAA500 has high detectable joints~\cite{alphapose} of 69.7\%, well above other representative action datasets.

\begin{table}
\begin{center}
{\small 
\begin{tabular}{r|c}
\hline 
Dataset                         & Detectable Joints            \\ \hline
Kinetics 400~\cite{kinetics400} &         41.0\%               \\ \hline
UCF101~\cite{ucf101}            &         37.8\%               \\ \hline
HMDB51~\cite{HMDB51}            &         41.8\%               \\ \hline
FineGym~\cite{finegym}          &         44.7\%               \\ \hline
\textbf{HAA500}                 & \textbf{69.7\%}              \\ \hline
\end{tabular}
}
\end{center}
\caption{Detectable joints of video action datasets. We use AlphaPose~\cite{alphapose} to detect the largest person in the frame, and count the number of joints with a score higher than $0.5$. }
\label{table:comparison_joint}
\vspace{-1.5em}
\end{table}

\begin{table*}[hbt!]
\begin{center}
\minipage[t]{0.347 \linewidth}
{\small 
        \begin{tabular}{c|c|c|c}
            \hline
            \multicolumn{2}{c|}{} &
            \multicolumn{2}{c}{500 Atomic} \\
            \hline
              \multicolumn{2}{c|}{Model}   & Top-1 & Top-3  \\
            \hline
            \multirow{4}{*}{I3D~\cite{i3d}} & RGB & 33.53\% & 53.00\%   \\
            & Flow                                & 34.73\% & 52.40\% \\
            & Pose                                & 35.73\% & 54.07\%  \\
            & Three-Stream                        & 49.87\% & 66.60\%  \\
            \hline
            \multirow{4}{*}{SlowFast~\cite{slowfast}} & RGB & 25.07\% & 44.07\%  \\
            & Flow                                          & 22.87\% & 36.93\%  \\
            & Pose                                          & 28.33\% & 45.20\%  \\
            & Three-Stream                                  & 39.93\% & 56.00\%  \\
            \hline
            \multirow{3}{*}{TSN~\cite{TSN}} & RGB & 55.33\% & 75.00\% \\
                                 & Flow & 49.13\% & 66.60\% \\
                                 & Two-Stream & 64.40\% & 80.13\% \\
            \hline
            \multirow{1}{*}{TPN~\cite{TPN}} & RGB & 50.53\% & 68.13\% \\
            \hline
            \multirow{1}{*}{ST-GCN~\cite{stgcn}} & Pose & 29.67\% & 47.13\% \\
            \hline
        \end{tabular}
}
\endminipage
~~~~~
\minipage[t]{0.205 \linewidth}
{\small 
        \begin{tabular}{c|c}
            \hline
            \multicolumn{1}{c|}{Inst.} &
            \multicolumn{1}{c}{Inst. with Atomic} \\
            \hline
            Top-1 & Top-1  \\
            \hline
             70.59\% & \bf{71.90}\%  \\
             73.20\% & \bf{77.79}\%  \\
             69.28\% & \bf{71.90}\%  \\
             81.70\% &     82.35\%   \\
            \hline
             40.52\% & \bf{50.98}\%  \\
             71.90\% &     71.90 \%  \\
             64.71\% & \bf{66.01}\%  \\
             67.97\% & \bf{73.86}\%  \\
            \hline
             \bf{86.93}\% & 84.31\%  \\
             79.08\% & \bf{86.27}\%  \\
             89.54\% & 90.20\%  \\
            \hline
             73.20\% & \bf{75.82}\%  \\
            \hline
             67.32\% & 67.97\%  \\
             \hline
        \end{tabular}
}
\endminipage
~~~~~
\minipage[t]{0.205 \linewidth}
{\small 
        \begin{tabular}{c|c}
            \hline
            \multicolumn{1}{c|}{Sport} &
            \multicolumn{1}{c}{Sport with Atomic} \\
            \hline
            Top-1 & Top-1  \\
            \hline
             47.48\% & \bf{53.93}\%  \\
             51.42\% & \bf{54.40}\%  \\
             54.87\% &     55.03 \%  \\
             68.55\% & \bf{69.81}\%   \\
            \hline
             42.92\% & \bf{44.18}\%  \\
             44.81\% & \bf{45.91}\%  \\
             42.45\% & \bf{50.00}\%  \\
             59.91\% & \bf{62.89}\%  \\
            \hline
             72.64\% & 72.48\%  \\
             \bf{69.97}\% & 68.24\%  \\
             \bf{81.13}\% & 78.93\%  \\
            \hline
             61.64\% & \bf{64.15}\%  \\
            \hline
             40.25\% & \bf{43.87}\%  \\
             \hline
        \end{tabular}
}
\endminipage
\end{center}
\caption{ \textbf{Left}: HAA500 trained over different models. \textbf{Right}: Composite action classification accuracy of different models when they are trained with/without atomic action classification. Numbers are bolded when the difference is larger than 1\%. }
\vspace{-0.2in}
\label{table:Experiments}

\end{table*}
\vspace{-1em}
\subsubsection{Atomic}

\begin{figure}[!t]
\vspace{-0.5em}
\begin{center}
\includegraphics[width=\linewidth]{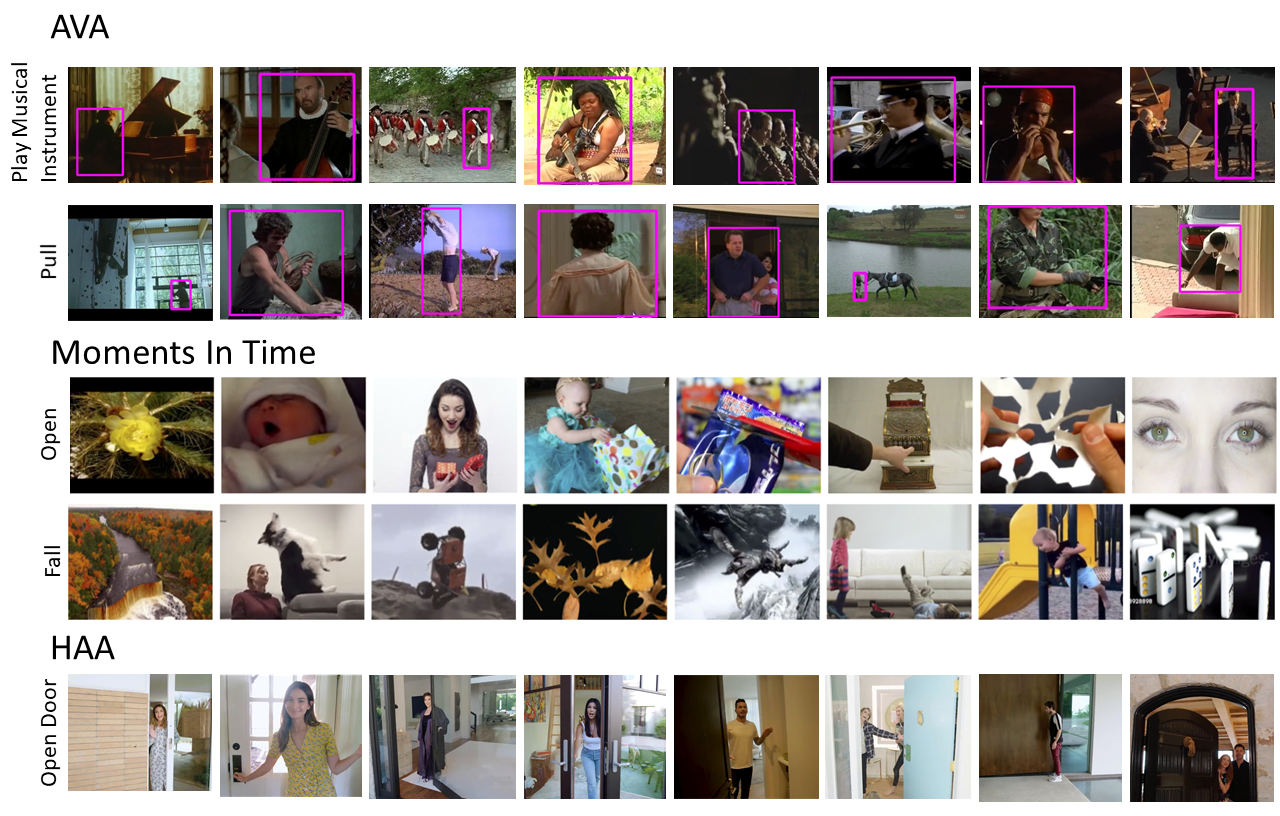}
\caption{Coarse-grained atomic action datasets label different actions under a single English action verb. HAA500 (Bottom) has fine-grained classes where the action ambiguities are eliminated as much as possible.}
\label{fig:comparison_atomicity}
\end{center}
\vspace{-2em}
\end{figure}

Existing atomic action datasets such as \cite{ava_speech,AVA,momentsintime} are limited by English linguistics, where action verbs (\eg, walk, throw, pull, \etc) are decomposed. Such classification does not fully eliminate the aforementioned problems of composite action datasets. Figure~\ref{fig:comparison_atomicity} shows cases of different atomic action datasets where a single action class contains fundamentally different actions. 

On the other hand, our fine-grained atomic actions contain only a single type of action under each class, \eg, \textit{Baseball - Pitch}, \textit{Yoga - Tree}, \textit{Hopscotch - Spin}, \etc 

\vspace{-1em}
\subsubsection{Scalability}
\vspace{-0.5em}

Requiring only 20 video annotations per class, or around 600 frames to characterize a human-centric atomic action curated as described above, our class-balanced dataset is highly scalable compared to other representative datasets requiring annotation of hundreds or even thousands of videos. In practice, our annotation per class takes around 20--60 minutes including searching the Internet for videos with expected quality. The detailed annotation procedure is available in the supplementary material.

\section{Empirical Studies}

We study HAA500 over multiple aspects using widely used action recognition models.
Left of Table~\ref{table:Experiments} shows the performance of the respective models when they are trained with HAA500.
For a fair comparison between different models and training datasets, all the experiments have been performed using hyper parameters given by the original authors without ImageNet~\cite{imagenet} pre-training.

For Pose models except for ST-GCN~\cite{stgcn}, we use three-channel pose joint heatmaps~\cite{alphapose} to train pose models. RGB, Flow~\cite{flownet2} and Pose~\cite{alphapose} all show relatively similar performance in HAA500, where none of them shows superior performance than the others.  Given that pose heatmap has far less information than  given from RGB frames or optical flow frames, we expect that easily detectable joints of HAA500  benefit the pose-based model performance. 

Furthermore, we study the benefits of atomic action annotation on video recognition, as well as  the importance of human-centric characteristics of HAA500.
In this paper, we use I3D-RGB~\cite{i3d} with 32 frames for all of our experiments unless otherwise specified. We use AlphaPose~\cite{alphapose} for the models that require human pose estimation.

\subsection{Visualization}

\begin{figure}[t!]
\begin{center}
\minipage[t]{0.49 \linewidth}
    \includegraphics[width=\linewidth]{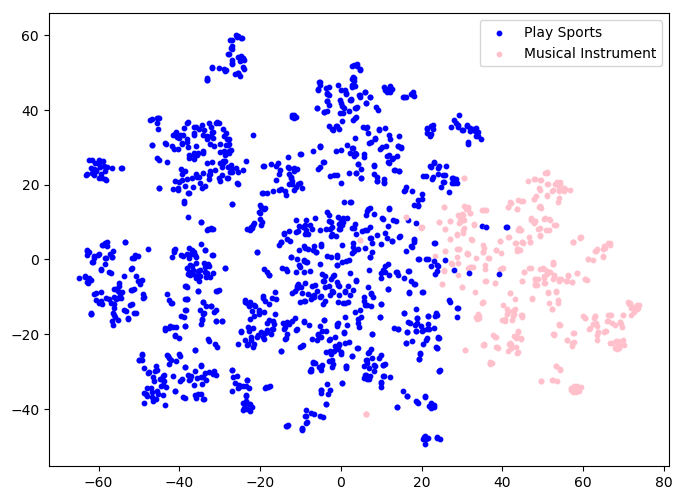}
\endminipage
\minipage[t]{0.49 \linewidth}
    \includegraphics[width=\linewidth]{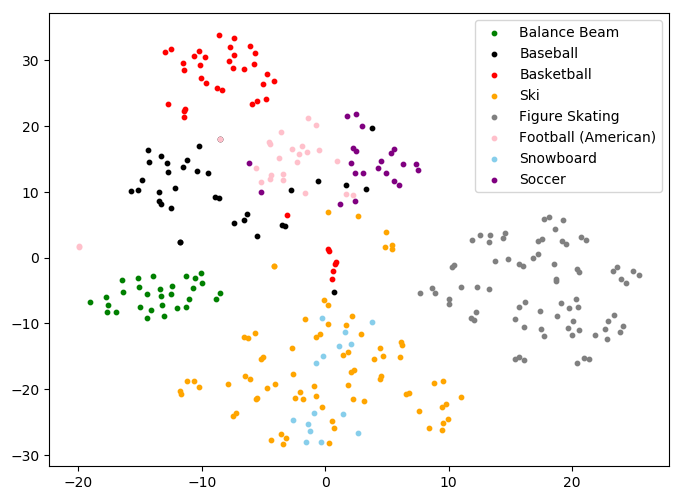}
\endminipage
\end{center}
\vspace{-1em}
\caption{Visualization of HAA500. We extract 1024-vectors from the second last layer of RGB-I3D and plot them using t-SNE.}
\label{fig:visualization}
\vspace{-1.3em}
\end{figure}

To study the atomic action recognition, we train RGB-I3D model on HAA500 and extract embedding vectors from the second last layer and plot them using truncated SVD and t-SNE. From Figure~\ref{fig:visualization}, the embedding vectors show clear similarities to the natural hierarchy of human action. On the left of the  figure, we see a clear distinction between classes in \textit{Playing Sports} and classes in \textit{Playing Musical Instruments}. Specifically, in sports, we see similar super-classes, \textit{Snowboarding} and \textit{Skiing}, under close embedding space, while \textit{Basketball}, \textit{Balance Beam (Gymnastics)}, and \textit{Figure Skating} are in their distinctive independent spaces. 
We observe super-class clustering of composite actions when only the atomic action labeling has been used to train the model. 
This visualization hints the benefit of fine-grained atomic action labeling for composite action classification tasks.

% ---------------------------------------------------------------
\vspace{-0.25em}
\subsection{Atomic Action}
\vspace{-0.25em}

We have previously discussed that modern action recognition datasets introduce ambiguities where two or more composite actions sharing the same atomic actions, while a single composite action class may contain multiple distinguishable actions (\eg, a composite action \textit{Playing Soccer} has \textit{Soccer-Dribble}, \textit{Soccer-Throw}, \etc). HAA500 addresses this issue by providing fine-grained atomic action labels that distinguish similar atomic action in different composite actions. 

To study the benefits of atomic action labels, specifically, how it helps composite action classification for ambiguous classes, we selected two areas from HAA500, \textit{Sports/Athletics} and \textit{Playing Musical Instruments}, in which composite actions contain strong ambiguities with other actions in the area. We compare models trained with two different types of labels: 1) only composite labels and 2) atomic + composite labels, then we evaluate the performance on composite action classification. Results are tabulated on the right of Table \ref{table:Experiments}. Accuracy of the models trained with only composite labels are under \textit{Inst.} and \textit{Sport} column, and the accuracy of composite action classification trained with atomic action classification is listed on the other columns. 

We can observe improvements in composite action classification when atomic action classification is incorporated. The fine-grained action decomposition in HAA500 enables the models to resolve ambiguities of similar atomic actions and helps the model to learn the subtle differences in the atomic actions across different composite actions. 
This demonstrates the importance of proper labeling of fine-grained atomic action which can increase the performance for composite action classification without changing the model architecture or the training set. 

% ---------------------------------------------------------------
% \vspace{-0.5em}
\subsection{Human-Centric}

HAA500 is designed to contain action clips with a high percentage of detectable human figures.
To study the importance of human-pose in fine-grained atomic action recognition, we compare the performance of HAA500 and FineGym when both RGB and pose estimation are given as input. For pose estimation, we obtain the 17 joint heatmaps from AlphaPose~\cite{alphapose} and merge them into 3 channels; head, upper-body, and lower-body.

\begin{table}[t]
    {\small 
    \begin{center}
        \begin{tabular}{l |c |c || c }
        \hline
          & RGB  & Pose & RGB + Pose \\
        \hline
        HAA500                    & 33.53\% & 35.73\% & 42.80\% \\
        \:\:\: Sport              & 38.52\% & 47.33\% & 50.94\% \\ 
        \:\:\: Instrument         & 30.72\% & 24.18\% & 32.03\% \\ 
        \:\:\: Hobbies            & 31.30\% & 26.42\% & 35.37\% \\ 
        \:\:\: Daily              & 28.82\% & 28.60\% & 39.14\% \\
        % \hline
        Gym288~\cite{finegym} & 76.11\% & 65.16\% & 77.31\% \\
        \hline
        \end{tabular}
    \end{center}}
    \caption{Atomic action classification accuracy when both RGB image and pose estimation are given as an input. We also show performance when they are trained separately for comparison.}
    \label{table:human-pose}
    \vspace{-0.5em}
\end{table}

Table~\ref{table:human-pose} tabulates the results. In three out of four areas of HAA500, I3D-RGB shows better performance than I3D-Pose, due to the vast amount of information given to the model. I3D-Pose shows the highest performance on \textit{Sports/Athletics} with vibrant and distinctive action, while I3D-Pose fails to show comparable performance in \textit{Playing Musical Instrument} area, where predicting the atomic action from only 17 joints is quite challenging.  Nonetheless, our experiments show a performance boost when both pose estimation and RGB frame are fed to the atomic action classification model, implicating the importance of human action in HAA500 action classification. For FineGym - Gym288, due to the rapid athletic movements resulting in blurred frames, the human pose is not easily recognizable which accounts for relatively insignificant improvements when pose has been used.

\section{Observations}
We present notable characteristics observed from HAA500 with our cross-dataset experiments.

\vspace{-1em}
\paragraph{Effects of Fine-Tuning over HAA500}
\begin{table}[t]
{\small 
\begin{center}
\begin{tabular}{l|c|c|c}
\hline
 & UCF101~\cite{ucf101} & ActNet 100~\cite{activitynet} & HMDB51~\cite{HMDB51} \\
\textbf{Pre-trained}          & Top-1            & Top-1            & Top-1 \\
\hline
None                          & 58.87\%          & 43.54\%          & 28.56\% \\ 
AVA~\cite{AVA}                & 48.54\%          & 30.51\%          & 25.28\% \\ 
Gym288~\cite{finegym}         & \textbf{69.94\%} & 43.79\%          & 36.24\% \\ 
UCF101~\cite{ucf101}          & -                & 42.94\%          & 32.37\% \\ 
ActNet 100~\cite{activitynet} & 57.52\%          & -                & 28.63\% \\
HMDB51~\cite{HMDB51}          & 53.36\%          & 39.33\%          &  -      \\
\hline
HAA500                        & 68.70\%          & \textbf{47.75\%} & \textbf{40.45\%}\\
\:\:\:Relaxed                & 62.24\%          & 38.30\%          & 33.29\%          \\ 
\hline
\end{tabular}
\end{center}}
\vspace{-0.5em}
\caption{Fine-tuning performance on I3D.}
\label{table:transfer}
\vspace{-0.8em}
\end{table}

Here, we test how to exploit the curated HAA500 dataset to detect action in ``in-the-wild" action datasets.  We pre-train I3D-RGB~\cite{i3d} using HAA500 or other video action datasets~\cite{activitynet,AVA,HMDB51,finegym,ucf101}, and freeze all the layers except for the last three for feature extraction. We then fine-tune the last three layers with ``in-the-wild" composite action datasets~\cite{activitynet,HMDB51,ucf101}.

Table~\ref{table:transfer} tabulates the fine-tuning result.
Our dataset is carefully curated to have a high variety of backgrounds and people while having consistent actions over each class. Despite being comparably smaller and more ``human-centric" than other action recognition datasets, HAA500's cleanness and high variety make it easily transferable to different tasks and datasets.

\vspace{-1em}
\paragraph{Effects of Scale Normalization} 

HAA500 has high diversity in human positions across the video collection. Here, we choose an area of HAA500, \textit{Playing Musical Instruments}, to investigate the effect of human-figure normalization on detection accuracy. We have manually annotated the bounding box of the person-of-interest in each frame and cropped them for the model to focus on the human action. In Table~\ref{table:normalize}, we test models that were trained to detect the composite actions or both composite and atomic actions. 

\begin{table}[t]
    {\small 
    \begin{center}
        \begin{tabular}{ l |c |c | c | c  }
        \hline
          & \multicolumn{2}{c|}{Original} & \multicolumn{2}{c}{Normalized} \\
          & Composite  & Both & Composite  & Both  \\
        \hline
        I3D-RGB  & 66.01\%  & 56.86\%          & \textbf{75.82\%} & \textbf{77.12\%} \\ 
        I3D-Flow & 73.20\%  & \textbf{77.78\%} & \textbf{75.16\%} & 74.51\% \\ 
        2-Stream & 77.78\%  & 80.39\%          & \textbf{83.01\%} & 80.39\% \\ 
        \hline
        \end{tabular}
    \end{center}}
    \vspace{-0.1in}
    \caption{Accuracy improvements on person-of-interest normalization. Numbers are composite action classification accuracy.}
    \label{table:normalize}
    \vspace{-0.2in}
\end{table}

While HAA500 is highly human-centric with person-of-interest as the most dominant figure of the frame, action classification on the normalized frames still shows considerable improvement when trained on either atomic action annotations or composite action annotations. This indicates the importance of spatial annotation for  action recognition.

\vspace{-0.5em}
\paragraph{Effects of Object Detection} 
In most video action datasets, non-human objects exist as a strong bias to the classes (\eg, basketball in \textit{Playing Basketball}). 
When highly diverse actions (\eg, \textit{Shooting a Basketball}, \textit{Dribbling a Basketball}, \etc) are annotated under a single class, straightforward deep-learning models tend to suffer from the bias and will learn to detect the easiest common factor (basketball) among the video clips, rather than ``seeing" the pertinent human action. Poorly designed video action dataset encourages the action classification model to trivially become an object detection model. 

In HAA500, every video clip in the same class contains compatible actions, making the common factor to be the ``action", while objects are regarded as ``ambiguities" that spread among different classes (\eg, basketball exists in both \textit{Shooting a Basketball} and \textit{Dribbling a Basketball}). To test the influence of ``object" in HAA500, we design an experiment similar to investigating the effect of human poses, as presented in Table~\ref{table:human-pose}, where we use object detection heatmap instead. Here we use Fast RCNN~\cite{fastrcnn}  trained with COCO~\cite{coco} dataset to generate the object heatmap. Among 80 detectable objects in COCO, we select 42 objects in 5 categories (sports equipment, food, animals, cutleries, and vehicles) to draw a 5-channel heatmap. Similar to Table~\ref{table:human-pose}, the heatmap channel is appended to the RGB channel as input. 

\begin{table}[t]
    {\small 
    \begin{center}
        \begin{tabular}{ l |c |c  }
        \hline
             & RGB & + Object\\
         \hline
            HAA500            & 33.53\% & 33.73\%  \\
            \:\:  Sport       & 38.52\% & 38.68\%  \\
            \:\:  Instrument  & 30.72\% & 30.07\%  \\
            \:\:  HAA-COCO    & 34.26\% & 34.26\%  \\
        \hline
            UCF101            & 57.65\% & 60.19\%   \\
        \hline
        \end{tabular}
    \end{center}}
    \vspace{-0.1in}
    \caption{Accuracy of I3D when trained with object heatmap. HAA-COCO denotes 147 classes of HAA500 expected to have objects that were detected.}
    \vspace{-0.1in}
    \label{table:object_detection}
\end{table}

Table~\ref{table:object_detection} tabulates the negligible effect of objects in atomic action classification of HAA500, including the classes that are expected to use the selected objects (HAA-COCO), while UCF101 shows improvements when object heatmap is used as a visual cue. 
Given the negligible effect of object heatmaps, 
we believe that fine-grained annotation of actions can effectively eliminate unwanted ambiguities or bias (``objects") while in UCF101 (composite action dataset), ``objects" can still affect  action prediction.

\vspace{-1.2em}
\paragraph{Effects of Dense Temporal Sampling}

The top of Table~\ref{table:action_oriented} tabulates the performance difference of HAA500 and other datasets over the number of frames used during training and testing. The bottom of Table~\ref{table:action_oriented} tabulates the performance with varying strides with a window size of 32 frames, except AVA which we test with 16 frames. Top-1 accuracies on action recognition are shown except AVA which shows mIOU due to its multi-labeled nature of the dataset. 

As expected, most datasets show the best performance when 32 frames are fed. AVA shows a drop in performance due to the irrelevant frames (\eg, action changes, camera cuts, \etc) included in the wider window. While all the datasets show comparable accuracy when the model only uses a single frame (\ie, when the problem has been reduced to a  ``Scene Recognition" problem), both HAA500 and Gym288 show a significant drop compared to their accuracy in 32 frames. While having an identical background contributes to the performance difference for Gym288, from HAA500, we see how temporal action movements are crucial for the detection of atomic actions, and they cannot be trivially detected using a simple scene detecting model. 

We also see that the density of the temporal window is another important factor in atomic action classification. We see that both HAA500 and Gym288, which are fine-grained action datasets, show larger performance drops when the frames have been sampled with strides of 2 or more, reflecting the importance of sampling for short temporal action movements in fine-grained action classification.

\begin{table}[t]
{\small 
\begin{center}
\begin{tabular}{c|c|c|c|c}
\hline
 \# of frames & HAA500 & UCF101~\cite{ucf101} & AVA~\cite{AVA} & Gym288~\cite{finegym} \\
\hline
1  & 19.93\% & 45.57\% & 33.57\% & 39.77\%\\
2  & 23.27\% & 47.26\% & 39.42\% & 44.68\%\\
4  & 24.40\% & 49.30\% & 39.48\% & 51.22\%\\
8  & 24.07\% & 49.80\% & 42.38\% & 59.64\%\\
16 & 28.20\% & 52.31\% & 43.11\% & 69.25\% \\
32 & 33.53\% & 57.65\% & 29.88\% & 76.11\% \\
\hline
stride 2 & 27.47\% & 57.23\%  & 41.49\% & 68.68\%\\
stride 4 & 23.87\% & 52.29\%  & 40.52\% & 60.76\%\\
stride 8 & 18.47\% & 47.95\%  & 38.45\% & 39.31\%\\
\hline
\end{tabular}
\end{center}}
\vspace{-0.1in}
\caption{Performance comparison on I3D-RGB over the number of frames and strides, wherein the latter a window size of 32 frames is used except AVA which we test with 16 frames.}
\label{table:action_oriented}
\vspace{-0in}
\end{table}

\vspace{-1em}
\paragraph{Quality versus Quantity}

\begin{table}[t]
    {\small 
    \begin{center}
        \begin{tabular}{l |c |c}
        \hline
          & HAA500  & Relaxed \\
        \hline
        Overall                   & \textbf{33.53\%} & 22.80\% \\
        \:\:\: Sport              & \textbf{38.52\%} & 25.47\% \\ 
        \:\:\: Instrument         & \textbf{30.72\%} & 28.10\% \\ 
        \:\:\: Hobbies            & \textbf{31.30\%} & 20.33\% \\ 
        \:\:\: Daily              & \textbf{28.82\%} & 18.71\% \\
        \hline
        \end{tabular}
    \end{center}
    \vspace{-0.1in}
    \caption{Action classification accuracy of original HAA500 and the relaxed version.}
    \label{table:dirty_basic}}
    \vspace{-0.20in}
\end{table}

To study the importance of our precise temporal annotation against the size of a dataset, we modify HAA500 by relaxing the temporal annotation requirement, \ie, we take a longer clip than the original annotation. 
Our relaxed-HAA500 consists of 4400K labeled frames, a significant increase from the original HAA500 with 591K frames. 
Table~\ref{table:dirty_basic} tabulates the performance comparison between the original and the relaxed version of HAA500 on the original HAA500 test set. 
We observe the performance drop in all areas, with a significant drop in \textit{Playing Sports}, where accurate temporal annotation benefits the most. Performance drop in \textit{Playing Musical Instruments} area is less significant, as start/finish of action is vaguely defined in these classes.
We also test the fine-tuning performance of relaxed-HAA500, where the bottom-most row of Table~\ref{table:transfer} tabulates the performance drop when the relaxed-HAA500 is used for pre-training. Both of our experiments show the importance of accurate temporal labeling over the size of a dataset.

\vspace{-0.5em}
\section{Conclusion}
\vspace{-0.5em}

This paper introduces HAA500, a new human action dataset with fine-grained atomic action labels and human-centric clip annotations, where the videos are carefully selected such that the relevant human poses are apparent and detectable. With carefully curated action videos, HAA500 does not suffer from irrelevant frames, where videos clips only exhibit the annotated action. With a small number of clips per class, HAA500 is highly scalable to include more action classes. We have demonstrated the efficacy of HAA500 where action recognition can be greatly benefited from our clean, highly diversified, class-balanced fine-grained atomic action dataset which is human-centric with a high percentage of detectable poses. On top of HAA500, we have also empirically investigated several important factors that can affect the performance of action recognition. We hope HAA500 and our findings could facilitate new advances in video action recognition.

{\small
\bibliographystyle{ieee_fullname}
\bibliography{main}

\begin{thebibliography}{10}\itemsep=-1pt

\bibitem{alwassel2018diagnosing}
Humam Alwassel, Fabian~Caba Heilbron, Victor Escorcia, and Bernard Ghanem.
\newblock Diagnosing error in temporal action detectors.
\newblock In {\em ECCV 2018}.

\bibitem{DBLP:conf/iccv/BlankGSIB05}
Moshe Blank, Lena Gorelick, Eli Shechtman, Michal Irani, and Ronen Basri.
\newblock Actions as space-time shapes.
\newblock In {\em ICCV 2005}.

\bibitem{kinetics700}
Joao Carreira, Eric Noland, Chloe Hillier, and Andrew Zisserman.
\newblock A short note on the kinetics-700 human action dataset.
\newblock {\em arXiv preprint arXiv:1907.06987}, 2019.

\bibitem{i3d}
Jo{\~{a}}o Carreira and Andrew Zisserman.
\newblock Quo vadis, action recognition? {A} new model and the kinetics
  dataset.
\newblock In {\em {CVPR 2017}}.

\bibitem{ava_speech}
Sourish Chaudhuri, Joseph Roth, Daniel P.~W. Ellis, Andrew~C. Gallagher, Liat
  Kaver, Rebecca Marvin, Caroline Pantofaru, Nathan Reale, Loretta~Guarino
  Reid, Kevin~W. Wilson, and Zhonghua Xi.
\newblock Ava-speech: A densely labeled dataset of speech activity in movies.
\newblock In {\em INTERSPEECH}, 2018.

\bibitem{epickitchens}
Dima Damen, Hazel Doughty, Giovanni Maria~Farinella, Sanja Fidler, Antonino
  Furnari, Evangelos Kazakos, Davide Moltisanti, Jonathan Munro, Toby Perrett,
  Will Price, et~al.
\newblock Scaling egocentric vision: The epic-kitchens dataset.
\newblock In {\em ECCV 2018}.

\bibitem{imagenet}
Jia Deng, Wei Dong, Richard Socher, Li-Jia Li, Kai Li, and Li Fei-Fei.
\newblock Imagenet: A large-scale hierarchical image database.
\newblock In {\em CVPR 2009}.

\bibitem{LRCN}
Jeff Donahue, Lisa~Anne Hendricks, Sergio Guadarrama, Marcus Rohrbach,
  Subhashini Venugopalan, Trevor Darrell, and Kate Saenko.
\newblock Long-term recurrent convolutional networks for visual recognition and
  description.
\newblock In {\em CVPR 2015}.

\bibitem{activitynet}
Bernard~Ghanem Fabian Caba~Heilbron, Victor~Escorcia and Juan~Carlos Niebles.
\newblock Activitynet: A large-scale video benchmark for human activity
  understanding.
\newblock In {\em CVPR 2015}.

\bibitem{alphapose}
Hao-Shu Fang, Shuqin Xie, Yu-Wing Tai, and Cewu Lu.
\newblock {RMPE}: Regional multi-person pose estimation.
\newblock In {\em ICCV 2017}.

\bibitem{slowfast}
Christoph Feichtenhofer, Haoqi Fan, Jitendra Malik, and Kaiming He.
\newblock Slowfast networks for video recognition.
\newblock In {\em {CVPR 2019}}.

\bibitem{feichtenhofer2016convolutional}
Christoph Feichtenhofer, Axel Pinz, and Andrew Zisserman.
\newblock Convolutional two-stream network fusion for video action recognition.
\newblock In {\em CVPR 2016}.

\bibitem{gaidon2013temporal}
Adrien Gaidon, Zaid Harchaoui, and Cordelia Schmid.
\newblock Temporal localization of actions with actoms.
\newblock {\em TPAMI 2013}.

\bibitem{jigsaws}
Yixin Gao, S~Swaroop Vedula, Carol~E Reiley, Narges Ahmidi, Balakrishnan
  Varadarajan, Henry~C Lin, Lingling Tao, Luca Zappella, Benjam{\i}n B{\'e}jar,
  David~D Yuh, et~al.
\newblock Jhu-isi gesture and skill assessment working set (jigsaws): A
  surgical activity dataset for human motion modeling.
\newblock In {\em Miccai workshop: M2cai 2014}.

\bibitem{fastrcnn}
Ross Girshick.
\newblock Fast r-cnn.
\newblock In {\em CVPR 2015}.

\bibitem{goyal2017something}
Raghav Goyal, Samira~Ebrahimi Kahou, Vincent Michalski, Joanna Materzynska,
  Susanne Westphal, Heuna Kim, Valentin Haenel, Ingo Fruend, Peter Yianilos,
  Moritz Mueller-Freitag, et~al.
\newblock The ``something something" video database for learning and evaluating
  visual common sense.
\newblock In {\em ICCV 2017}.

\bibitem{AVA}
Chunhui Gu, Chen Sun, David~A. Ross, Carl Vondrick, Caroline Pantofaru, Yeqing
  Li, Sudheendra Vijayanarasimhan, George Toderici, Susanna Ricco, Rahul
  Sukthankar, Cordelia Schmid, and Jitendra Malik.
\newblock {AVA:} {A} video dataset of spatio-temporally localized atomic visual
  actions.
\newblock In {\em CVPR 2018}.

\bibitem{flownet2}
Eddy Ilg, Nikolaus Mayer, Tonmoy Saikia, Margret Keuper, Alexey Dosovitskiy,
  and Thomas Brox.
\newblock Flownet 2.0: Evolution of optical flow estimation with deep networks.
\newblock In {\em CVPR 2017}.

\bibitem{DBLP:conf/iccv/JhuangGZSB13}
Hueihan Jhuang, Juergen Gall, Silvia Zuffi, Cordelia Schmid, and Michael~J.
  Black.
\newblock Towards understanding action recognition.
\newblock In {\em ICCV 2013}.

\bibitem{stm}
Boyuan Jiang, MengMeng Wang, Weihao Gan, Wei Wu, and Junjie Yan.
\newblock {STM}: Spatiotemporal and motion encoding for action recognition.
\newblock In {\em ICCV 2019}.

\bibitem{kinetics400}
Will Kay, Joao Carreira, Karen Simonyan, Brian Zhang, Chloe Hillier, Sudheendra
  Vijayanarasimhan, Fabio Viola, Tim Green, Trevor Back, Paul Natsev, et~al.
\newblock The kinetics human action video dataset.
\newblock {\em arXiv preprint arXiv:1705.06950}, 2017.

\bibitem{ke2017new}
Qiuhong Ke, Mohammed Bennamoun, Senjian An, Ferdous Sohel, and Farid Boussaid.
\newblock A new representation of skeleton sequences for 3d action recognition.
\newblock In {\em CVPR 2017}.

\bibitem{kim2017interpretable}
Tae~Soo Kim and Austin Reiter.
\newblock Interpretable 3d human action analysis with temporal convolutional
  networks.
\newblock In {\em CVPRW 2017}.

\bibitem{breakfast}
Hilde Kuehne, Ali Arslan, and Thomas Serre.
\newblock The language of actions: Recovering the syntax and semantics of
  goal-directed human activities.
\newblock In {\em CVPR 2014}.

\bibitem{HMDB51}
Hildegard Kuehne, Hueihan Jhuang, Est{\'{\i}}baliz Garrote, Tomaso~A. Poggio,
  and Thomas Serre.
\newblock {HMDB:} {A} large video database for human motion recognition.
\newblock In {\em ICCV 2011}.

\bibitem{diving48}
Yingwei Li, Yi Li, and Nuno Vasconcelos.
\newblock Resound: Towards action recognition without representation bias.
\newblock In {\em ECCV 2018}.

\bibitem{TSM}
Ji Lin, Chuang Gan, and Song Han.
\newblock {TSM}: Temporal shift module for efficient video understanding.
\newblock In {\em ICCV 2019}.

\bibitem{coco}
Tsung-Yi Lin, Michael Maire, Serge Belongie, James Hays, Pietro Perona, Deva
  Ramanan, Piotr Doll{\'a}r, and C.~Lawrence Zitnick.
\newblock Microsoft coco: Common objects in context.
\newblock In {\em ECCV 2014}.

\bibitem{DBLP:conf/cvpr/MarszalekLS09}
Marcin Marszalek, Ivan Laptev, and Cordelia Schmid.
\newblock Actions in context.
\newblock In {\em {CVPR} 2009}.

\bibitem{DBLP:conf/eccv/MettesGS16}
Pascal Mettes, Jan~C. van Gemert, and Cees G.~M. Snoek.
\newblock Spot on: Action localization from pointly-supervised proposals.
\newblock In {\em {ECCV} 2016}.

\bibitem{moltisanti2017trespassing}
Davide Moltisanti, Michael Wray, Walterio Mayol-Cuevas, and Dima Damen.
\newblock Trespassing the boundaries: Labeling temporal bounds for object
  interactions in egocentric video.
\newblock In {\em ICCV 2017}.

\bibitem{momentsintime}
Mathew Monfort, Alex Andonian, Bolei Zhou, Kandan Ramakrishnan, Sarah~Adel
  Bargal, Tom Yan, Lisa Brown, Quanfu Fan, Dan Gutfruend, Carl Vondrick, et~al.
\newblock Moments in time dataset: one million videos for event understanding.
\newblock {\em TPAMI 2019}.

\bibitem{acarnet}
Junting Pan, Siyu Chen, Mike~Zheng Shou, Yu Liu, Jing Shao, and Hongsheng Li.
\newblock Actor-context-actor relation network for spatio-temporal action
  localization.
\newblock In {\em CVPR 2021}.

\bibitem{DBLP:conf/cvpr/RodriguezAS08}
Mikel~D. Rodriguez, Javed Ahmed, and Mubarak Shah.
\newblock Action {MACH} a spatio-temporal maximum average correlation height
  filter for action recognition.
\newblock In {\em CVPR 2008}.

\bibitem{MPIICooking2}
Marcus Rohrbach, Anna Rohrbach, Michaela Regneri, Sikandar Amin, Mykhaylo
  Andriluka, Manfred Pinkal, and Bernt Schiele.
\newblock Recognizing fine-grained and composite activities using hand-centric
  features and script data.
\newblock {\em IJCV 2016}.

\bibitem{ava_speaker}
Joseph Roth, Sourish Chaudhuri, Ondrej Klejch, Radhika Marvin, Andrew
  Gallagher, Liat Kaver, Sharadh Ramaswamy, Arkadiusz Stopczynski, Cordelia
  Schmid, Zhonghua Xi, et~al.
\newblock Ava active speaker: An audio-visual dataset for active speaker
  detection.
\newblock In {\em 2020 IEEE International Conference on Acoustics, Speech and
  Signal Processing}.

\bibitem{DBLP:conf/icpr/SchuldtLC04}
Christian Sch{\"{u}}ldt, Ivan Laptev, and Barbara Caputo.
\newblock Recognizing human actions: {A} local {SVM} approach.
\newblock In {\em CVPR 2004}.

\bibitem{nturgbd}
Amir Shahroudy, Jun Liu, Tian-Tsong Ng, and Gang Wang.
\newblock {NTU RGB+D}: A large scale dataset for 3d human activity analysis.
\newblock In {\em CVPR 2016}.

\bibitem{finegym}
Dian Shao, Yue Zhao, Bo Dai, and Dahua Lin.
\newblock Finegym: A hierarchical video dataset for fine-grained action
  understanding.
\newblock In {\em CVPR 2020}.

\bibitem{sigurdsson2016hollywood}
Gunnar~A. Sigurdsson, G{\"u}l Varol, Xiaolong Wang, Ali Farhadi, Ivan Laptev,
  and Abhinav Gupta.
\newblock Hollywood in homes: Crowdsourcing data collection for activity
  understanding.
\newblock In {\em ECCV 2016}.

\bibitem{DBLP:conf/nips/SimonyanZ14}
Karen Simonyan and Andrew Zisserman.
\newblock Two-stream convolutional networks for action recognition in videos.
\newblock In {\em NIPS 2014}.

\bibitem{ucf101}
Khurram Soomro, Amir~Roshan Zamir, and Mubarak Shah.
\newblock Ucf101: A dataset of 101 human actions classes from videos in the
  wild.
\newblock {\em arXiv preprint arXiv:1212.0402}, 2012.

\bibitem{TSN}
Limin Wang, Yuanjun Xiong, Zhe Wang, Yu Qiao, Dahua Lin, Xiaoou Tang, and Luc
  Van~Gool.
\newblock Temporal segment networks for action recognition in videos.
\newblock {\em TPAMI 2018}.

\bibitem{DBLP:journals/corr/WeinzaepfelMS16}
Philippe Weinzaepfel, Xavier Martin, and Cordelia Schmid.
\newblock Towards weakly-supervised action localization.
\newblock {\em arXiv preprint arXiv:1605.05197}, 2, 2016.

\bibitem{stgcn}
Sijie Yan, Yuanjun Xiong, and Dahua Lin.
\newblock Spatial temporal graph convolutional networks for skeleton-based
  action recognition.
\newblock In {\em AAAI 2018}.

\bibitem{TPN}
Ceyuan Yang, Yinghao Xu, Jianping Shi, Bo Dai, and Bolei Zhou.
\newblock Temporal pyramid network for action recognition.
\newblock In {\em CVPR 2020}.

\bibitem{DBLP:conf/cvpr/YuanLW09}
Junsong Yuan, Zicheng Liu, and Ying Wu.
\newblock Discriminative subvolume search for efficient action detection.
\newblock In {\em CVPR 2009}.

\bibitem{zhao2019hacs}
Hang Zhao, Antonio Torralba, Lorenzo Torresani, and Zhicheng Yan.
\newblock Hacs: Human action clips and segments dataset for recognition and
  temporal localization.
\newblock In {\em ICCV 2019}.

\bibitem{TRN}
Bolei Zhou, Alex Andonian, Aude Oliva, and Antonio Torralba.
\newblock Temporal relational reasoning in videos.
\newblock In {\em ECCV 2018}.

\end{thebibliography}
}

\end{document}

%%%%%%%%% TITLE
\title{HAA500: Supplementary Material}

\author{
\begin{tabular}{ccccc}
Jihoon Chung$^{1,2}$ & Cheng-hsin Wuu$^{1,3}$ & Hsuan-ru Yang$^1$ & Yu-Wing Tai$^{1,4}$ & Chi-Keung Tang$^1$
\end{tabular}
\\
$^1$HKUST $^2$ Princeton University $^3$Carnegie Mellon University $^4$Kuaishou Technology
\\
{\tt\small jc5933@princeton.edu cwuu@andrew.cmu.edu hyangap@ust.hk yuwing@gmail.com cktang@cs.ust.hk}
% To save space, use either the email address or home page, not both\
}

\maketitle
%\thispagestyle{empty}

%%%%%%%%% BODY TEXT
\section{Video Collection Procedure}

To guarantee a clean dataset with no label noises, we adopt a strict video collecting methodology for every class. We detail the method below.

\begin{enumerate}[itemsep=-2mm]
   \item We assign a single annotator for a single class. This is to assure that the same rule applies to every video in a class.
   \item The action class is classified as either continuous action or discrete action. Discrete action is when the action can have a single distinguishable action sequence. (\eg, \textit{Baseball-Swing}, \textit{Yoga-Bridge}, etc.). Continuous action otherwise. (\textit{Running}, \textit{Playing Violin}, etc.)
   \begin{enumerate}
     \item If it is discrete, make an internal rule to define the action. (e.g., \textit{Jumping Jack} starts and ends when the person is standing still. The video clip contains only a single jump. \textit{Push-up} starts and ends when the person is at the highest point. It should only have a single push-up). Every video should follow the internal rule so that every action in the class has compatible motion. 
     \item For continuous, we take video clips with appropriate length.
   \end{enumerate}
   
   \item Here are rules that the annotator has to follow.
   \begin{itemize}
       \item 20 videos should be unique to each other with a varied person, varied backgrounds.
       \item The person in action should be the dominant person of the frame. If there are people of non-interest, they should not be performing any action.
       \item Camera cuts should not exist.
       \item Every video should include a large portion of the human body.
       \item It is fine to have action variance that doesn't influence the semantics of the action. (\eg, a person can sit or stand in \textit{Whistling with One Hand} as long as the motion of whistling exists.)
       \item 20 videos are split into train/val/test set by 16/1/3. The validation set contains the ``standard" body action of the class, and 3 videos in the test set should be well diverse. 
   \end{itemize}
   \item Two or more reviewers that are not the annotator review the video to check for any mistakes. 
\end{enumerate}

\section{Experiment Detail}
In this section, we explain some of the experiment details of our paper. 

\paragraph{Variable Length of a Video}
For model~\cite{i3d,slowfast,TSN,TPN}, we randomly select 32 adjacent frames of a video during training. If the video is shorter than 32 frames, we replicate the last frame to match the size. During testing, we replicate the last frame to match the size to a multiple of 32, where the video is then divided into smaller mini-clips of size 32. The prediction score of each mini-clip is averaged to get the final prediction. In Table 11, where we train with fewer frames, we zero-pad on both ends to size 16. On ST-GCN~\cite{stgcn} we follow the same procedure of the original paper, where the video is either truncated or replicated to match the length of 300.

\paragraph{Implementation}
In all of our experiments, we use PyTorch for our deep learning framework. We use the official code of the model when they are available. While we use the same hyperparameters which the authors used for their model, for a fair comparison we do not pre-train the model before training.

\begin{figure*}[t!]
\begin{center}
    \vspace{-1em}
	\input{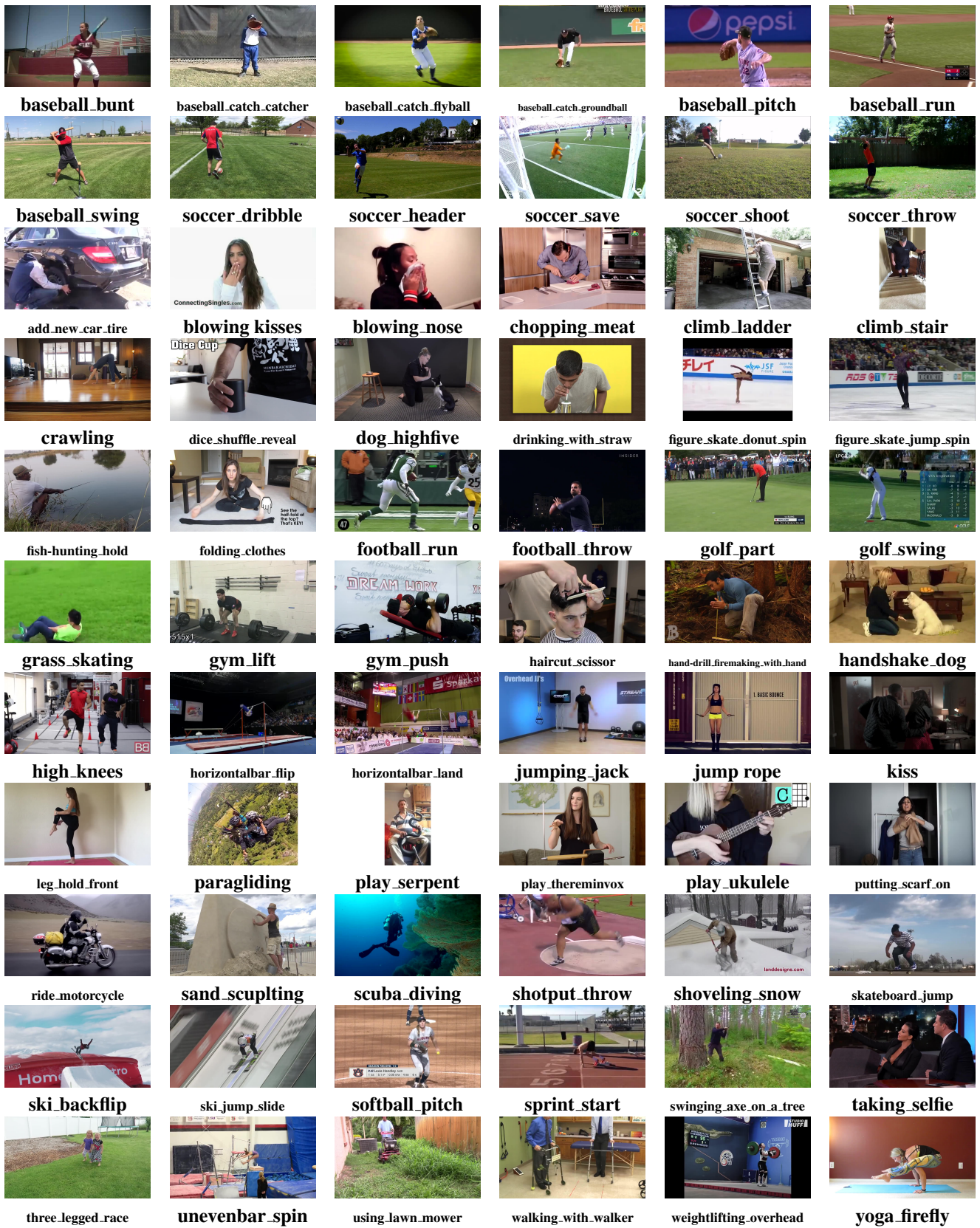}
\end{center}
\caption{Video samples of different classes.}
\label{fig:supp_0}
\end{figure*}

\begin{figure*}[t!]
\begin{center}
    \vspace{-1em}
	\input{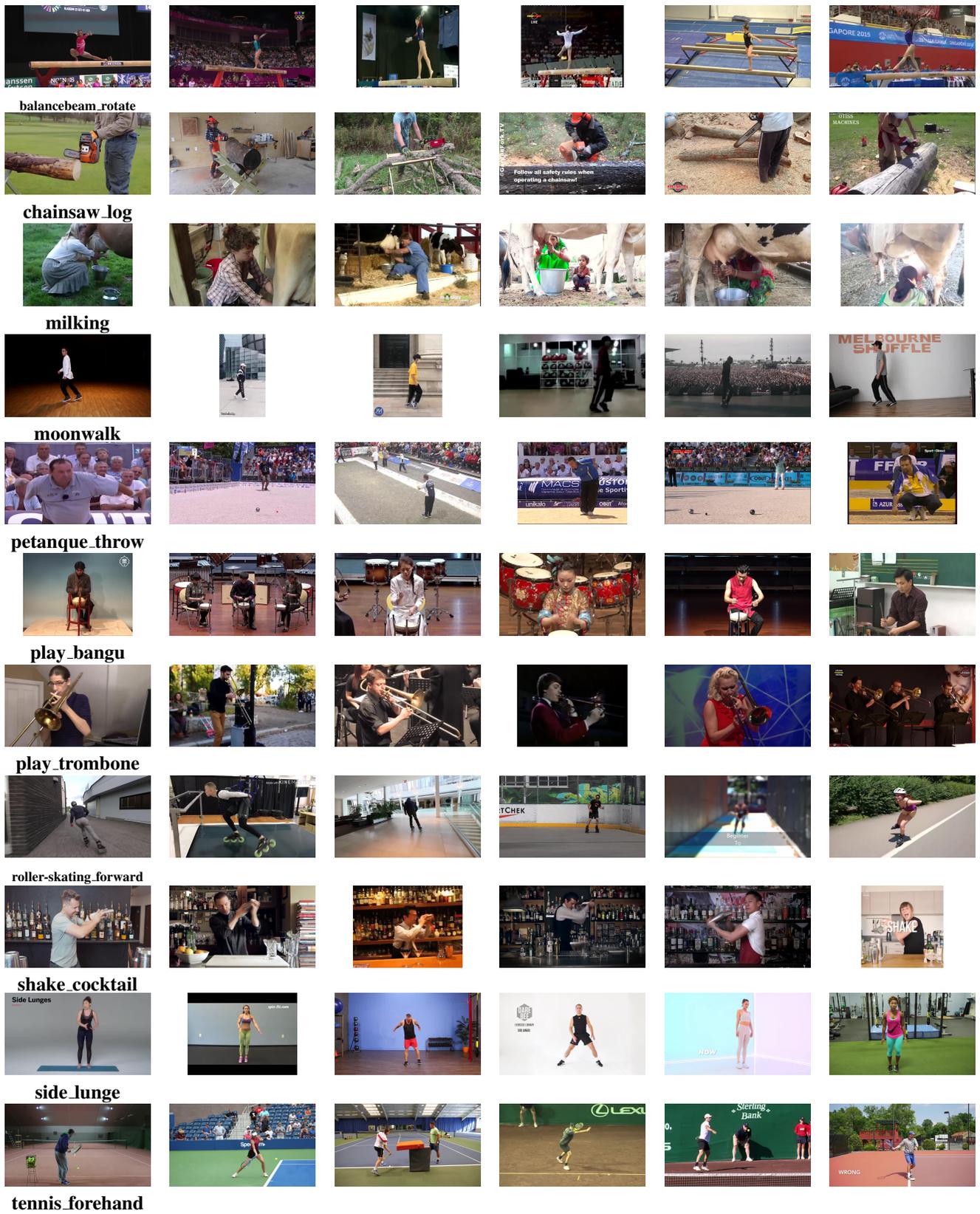}
\end{center}
\caption{HAA500 contains diverse videos per action class.}
\label{fig:supp_2}
\end{figure*}

\begin{figure*}[t!]
\begin{center}
    \vspace{-1em}
	\input{suppmat_figure2}
\end{center}
\caption{
Six sample frames of different videos. Each frame has an equal distance from the other, the first and the last sample frame are the first and the last frame of the video. In discrete action classes, (\textit{Long Jump - Jump}, \textit{Push Up}, \textit{Soccer - Shoot}), every video in the class shows a single motion. For action classes where it is hard to define a single motion (\ie, continuous actions, \eg, \textit{Play Violin}), videos are cut in appropriate length. 
}
\label{fig:supp_3}
\end{figure*}

\begin{figure*}[t!]
\begin{center}
    \vspace{-1em}
    \includegraphics[width=\linewidth]{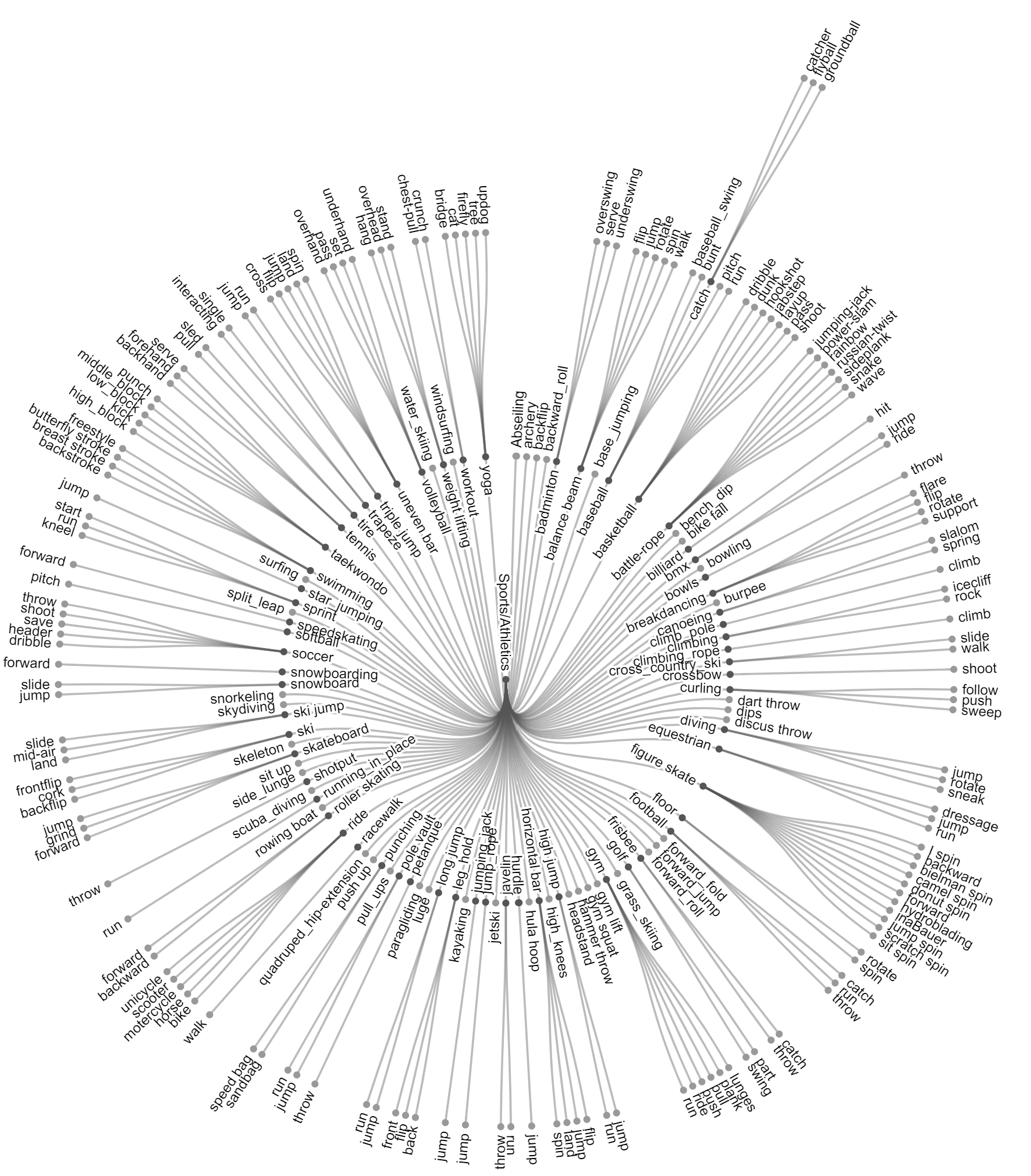}
\end{center}
\caption{Hierarchy of action classes in Sports/Athletics area.}
\label{fig:supp_1}
\end{figure*}

\section{List of Classes in HAA500}
Here, we list classes of HAA500 in each area. 

\paragraph{Sports/Athletics}
\begin{enumerate}[itemsep=-2mm]
    \item Abseiling
    \item Archery
    \item Backflip
    \item Backward Roll
    \item Badminton Overswing
    \item Badminton Serve
    \item Badminton Underswing
    \item Balance Beam Flip
    \item Balance Beam Jump
    \item Balance Beam Rotate
    \item Balance Beam Spin
    \item Balance Beam Walk
    \item Base Jumping
    \item Baseball Baseball Swing
    \item Baseball Bunt
    \item Baseball Pitch
    \item Baseball Run
    \item Basketball Dribble
    \item Basketball Dunk
    \item Basketball Hookshot
    \item Basketball Jabstep
    \item Basketball Layup
    \item Basketball Pass
    \item Basketball Shoot
    \item Battle-Rope Jumping-Jack
    \item Battle-Rope Power-Slam
    \item Battle-Rope Rainbow
    \item Battle-Rope Russian-Twist
    \item Battle-Rope Sideplank
    \item Battle-Rope Snake
    \item Battle-Rope Wave
    \item Bench Dip
    \item Bike Fall
    \item Billiard Hit
    \item Bmx Jump
    \item Bmx Ride
    \item Bowling
    \item Bowls Throw
    \item Breakdancing Flare
    \item Breakdancing Flip
    \item Breakdancing Rotate
    \item Breakdancing Support
    \item Burpee
    \item Canoeing Slalom
    \item Canoeing Spring
    \item Catch Catcher
    \item Catch Flyball
    \item Catch Groundball
    \item Climb Pole Climb
    \item Climbing Icecliff
    \item Climbing Rock
    \item Climbing Rope Climb
    \item Cross Country Ski Slide
    \item Cross Country Ski Walk
    \item Crossbow Shoot
    \item Curling Follow
    \item Curling Push
    \item Curling Sweep
    \item Dart Throw
    \item Dips
    \item Discus Throw
    \item Diving Jump
    \item Diving Rotate
    \item Diving Sneak
    \item Equestrian Dressage
    \item Equestrian Jump
    \item Equestrian Run
    \item Figure Skate I Spin
    \item Figure Skate Backward
    \item Figure Skate Bielman Spin
    \item Figure Skate Camel Spin
    \item Figure Skate Donut Spin
    \item Figure Skate Forward
    \item Figure Skate Hydroblading
    \item Figure Skate Inabauer
    \item Figure Skate Jump Spin
    \item Figure Skate Scratch Spin
    \item Figure Skate Sit Spin
    \item Floor Rotate
    \item Floor Spin
    \item Football Catch
    \item Football Run
    \item Football Throw
    \item Forward Fold
    \item Forward Jump
    \item Forward Roll
    \item Frisbee Catch
    \item Frisbee Throw
    \item Golf Part
    \item Golf Swing
    \item Grass Skiing
    \item Gym Lift
    \item Gym Lunges
    \item Gym Plank
    \item Gym Pull
    \item Gym Push
    \item Gym Ride
    \item Gym Run
    \item Gym Squat
    \item Hammer Throw
    \item Headstand
    \item High Jump Jump
    \item High Jump Run
    \item High Knees
    \item Horizontal Bar Flip
    \item Horizontal Bar Jump
    \item Horizontal Bar Land
    \item Horizontal Bar Spin
    \item Hula Hoop
    \item Hurdle Jump
    \item Javelin Run
    \item Javelin Throw
    \item Jetski
    \item Jump Rope Jump
    \item Jumping Jack Jump
    \item Kayaking
    \item Leg Hold Back
    \item Leg Hold Flip
    \item Leg Hold Front
    \item Long Jump Jump
    \item Long Jump Run
    \item Luge
    \item Paragliding
    \item Petanque Throw
    \item Pole Vault Jump
    \item Pole Vault Run
    \item Pull Ups
    \item Punching Sandbag
    \item Punching Speed Bag
    \item Push Up
    \item Quadruped Hip-Extension
    \item Racewalk Walk
    \item Ride Bike
    \item Ride Horse
    \item Ride Motercycle
    \item Ride Scooter
    \item Ride Unicycle
    \item Roller Skating Backward
    \item Roller Skating Forward
    \item Rowing Boat
    \item Running In Place Run
    \item Scuba Diving
    \item Shotput Throw
    \item Side Lunge
    \item Sit Up
    \item Skateboard Forward
    \item Skateboard Grind
    \item Skateboard Jump
    \item Skeleton
    \item Ski Backflip
    \item Ski Cork
    \item Ski Frontflip
    \item Ski Jump Land
    \item Ski Jump Mid-Air
    \item Ski Jump Slide
    \item Skydiving
    \item Snorkeling
    \item Snowboard Jump
    \item Snowboard Slide
    \item Snowboarding Forward
    \item Soccer Dribble
    \item Soccer Header
    \item Soccer Save
    \item Soccer Shoot
    \item Soccer Throw
    \item Softball Pitch
    \item Speedskating Forward
    \item Split Leap
    \item Sprint Kneel
    \item Sprint Run
    \item Sprint Start
    \item Star Jumping Jump
    \item Surfing
    \item Swimming Backstroke
    \item Swimming Breast Stroke
    \item Swimming Butterfly Stroke
    \item Swimming Freestyle
    \item Taekwondo High Block
    \item Taekwondo Kick
    \item Taekwondo Low Block
    \item Taekwondo Middle Block
    \item Taekwondo Punch
    \item Tennis Backhand
    \item Tennis Forehand
    \item Tennis Serve
    \item Tire Pull
    \item Tire Sled
    \item Trapeze Interacting
    \item Trapeze Single
    \item Triple Jump Jump
    \item Triple Jump Run
    \item Uneven Bar Cross
    \item Uneven Bar Flip
    \item Uneven Bar Jump
    \item Uneven Bar Land
    \item Uneven Bar Spin
    \item Volleyball Overhand
    \item Volleyball Pass
    \item Volleyball Set
    \item Volleyball Underhand
    \item Water Skiing
    \item Weight Lifting Hang
    \item Weight Lifting Overhead
    \item Weight Lifting Stand
    \item Windsurfing
    \item Workout Chest-Pull
    \item Workout Crunch
    \item Yoga Bridge
    \item Yoga Cat
    \item Yoga Firefly
    \item Yoga Tree
    \item Yoga Updog
\end{enumerate}
\paragraph{Daily Actions}
\begin{enumerate}[itemsep=-2mm]
  \setcounter{enumi}{212}
    \item Add New Car Tire
    \item Adjusting Glasses
    \item ALS Icebucket Challenge
    \item Answering Questions
    \item Applauding
    \item Applying Cream
    \item Arm Wave
    \item Bandaging
    \item Bending Back
    \item Blowdrying Hair
    \item Blowing Balloon
    \item Blowing Glass
    \item Blowing Gum
    \item Blowing Kisses
    \item Blowing Leaf
    \item Blowing Nose
    \item Bowing Fullbody
    \item Bowing Waist
    \item Brushing Floor
    \item Brushing Hair
    \item Brushing Teeth
    \item Burping
    \item Calfrope Catch
    \item Calfrope Rope
    \item Calfrope Subdue
    \item Carrying With Head
    \item Cartwheeling
    \item Cast Net
    \item Chainsaw Log
    \item Chainsaw Tree
    \item Chalkboard
    \item Chewing Gum
    \item Chopping Meat
    \item Chopping Wood
    \item Cleaning Mirror
    \item Cleaning Mopping
    \item Cleaning Sweeping
    \item Cleaning Vacumming
    \item Cleaning Windows
    \item Clear Snow Off Car
    \item Climb Ladder
    \item Climb Stair
    \item Climbing Tree
    \item Closing Door
    \item CPR
    \item Crawling
    \item Cross Body Shoulder Stretch
    \item Cutting Onion
    \item Dabbing
    \item Dog Highfive
    \item Dog Walking
    \item Drinking With Cup
    \item Drinking With Straw
    \item Eat Apple
    \item Eat Burger
    \item Eat Spagetti
    \item Eating Hotdogs
    \item Eating Ice Cream
    \item Eating Oyster
    \item Face Slapping
    \item Falling Off Chair
    \item Fire Extinguisher
    \item Fist Bump
    \item Flamethrower
    \item Folding Blanket
    \item Folding Clothes
    \item Gas Pumping To Car
    \item Guitar Smashing
    \item Hailing Taxi
    \item Haircut Scissor
    \item Hammering Nail
    \item Hand In Hand
    \item Hand-Drill Firemaking Blow
    \item Hand-Drill Firemaking Drill With Bow
    \item Hand-Drill Firemaking Drill With Hand
    \item Handsaw
    \item Handshake Dog
    \item Hanging Clothes
    \item Headbang
    \item Heimlich Maneuver
    \item High Five
    \item Hold Baby
    \item Hold Baby With Wrap
    \item Hookah
    \item Hugging Animal
    \item Hugging Human
    \item Ironing Clothes
    \item Jack Up Car
    \item Kick Open Door
    \item Kiss
    \item Leaf Blowing
    \item Milking
    \item Neck Side Pull Stretch
    \item Opening Door
    \item Pancake Flip
    \item Peeling Banana
    \item Pizza Dough Toss
    \item Plunging Toilet
    \item Pottery Wheel
    \item Pouring Wine
    \item Push Car
    \item Push Wheelchair
    \item Push Wheelchair Alone
    \item Putting Scarf On
    \item Read Newspaper
    \item Reading Book
    \item Remove Car Tire
    \item Rescue Breathing
    \item Riding Camel
    \item Riding Elephant
    \item Riding Mechanical Bull
    \item Riding Mule
    \item Riding Ostrich
    \item Riding Zebra
    \item Rolling Snow
    \item Salute
    \item Screw Car Tire
    \item Setup Tent
    \item Shake Cocktail
    \item Shaking Head
    \item Shaving Beard
    \item Shoe Shining
    \item Shoveling Snow
    \item Sledgehammer Strike Down
    \item Smoking Exhale
    \item Smoking Inhale
    \item Spitting On Face
    \item Spraying Wall
    \item Sticking Tongue Out
    \item Stomping Grapes
    \item Styling Hair
    \item Swinging Axe On A Tree
    \item Talking Megaphone
    \item Talking On Phone
    \item Throwing Bouquet
    \item Using Inhaler
    \item Using Lawn Mower
    \item Using Lawn Mower Riding Type
    \item Using Metal Detector
    \item Using Scythe
    \item Using Spinning Wheel
    \item Using String Trimmer
    \item Using Typewriter
    \item Walking With Crutches
    \item Walking With Walker
    \item Wall Paint Brush
    \item Wall Paint Roller
    \item Washing Clothes
    \item Washing Dishes
    \item Watering Plants
    \item Wear Face Mask
    \item Wear Helmet
    \item Whipping
    \item Writing On Blackboard
    \item Yawning
\end{enumerate}
\paragraph{Musical Instruments}
\begin{enumerate}[itemsep=-2mm]
  \setcounter{enumi}{367}
    \item Accordian
    \item Bagpipes
    \item Bangu
    \item Banjo
    \item Bass Drum
    \item Bowsaw
    \item Cajon Drum
    \item Castanet
    \item Cello
    \item Clarinet
    \item Conga Drum
    \item Cornett
    \item Cymbals
    \item Doublebass
    \item Erhu
    \item Gong
    \item Grandpiano
    \item Guitar
    \item Handpan
    \item Harp
    \item Hulusi
    \item Jazzdrum
    \item Leaf-Flute
    \item Lute
    \item Maracas
    \item Melodic
    \item Noseflute
    \item Ocarina
    \item Otamatone
    \item Panpipe
    \item Piccolo
    \item Recorder
    \item Sanxian
    \item Saxophone
    \item Serpeng
    \item Sheng
    \item Sitar
    \item Snare Drum
    \item Sunoa
    \item Taiko Drum
    \item Tambourine
    \item Thereminvox
    \item Timpani
    \item Triangle
    \item Trombone
    \item Trumpet
    \item Ukulele
    \item Viola
    \item Violin
    \item Xylophone
    \item Yangqin
\end{enumerate}
\paragraph{Games and Hobbies}
\begin{enumerate}[itemsep=-2mm]
  \setcounter{enumi}{418}
    \item Air Drumming
    \item Air Guitar
    \item Air Hockey
    \item Alligator Wrestling
    \item Archaeological Excavation
    \item Arm Wrestling
    \item Atlatl Throw
    \item Axe Throwing
    \item Balloon Animal
    \item Beer Pong Throw
    \item Belly Dancing
    \item Blow Gun
    \item Building Snowman
    \item Card Throw
    \item Conducting
    \item Decorating Snowman
    \item Dice Shuffle Reveal
    \item Dice Stack Shuffle
    \item DJ
    \item Draw Handgun
    \item Face-Changing Opera
    \item Fire Breathing
    \item Fire Dancing Circulating
    \item Fish-Hunting Hold
    \item Fish-Hunting Pull
    \item Flipping Bottle
    \item Floss Dance
    \item Flying Kite
    \item Ganggangsullae
    \item Gangnam Style Dance
    \item Grass Skating
    \item Guitar Flip
    \item Hopscotch Pickup
    \item Hopscotch Skip
    \item Hopscotch Spin
    \item Ice Scuplting
    \item Juggling Balls
    \item Kick Jianzi
    \item Knitting
    \item Marble Scuplting
    \item Moonwalk
    \item Piggyback Ride
    \item Play Diabolo
    \item Play Kendama
    \item Play Yoyo
    \item Playing Nunchucks
    \item Playing Rubiks Cube
    \item Playing Seesaw
    \item Playing Swing
    \item Rock Balancing
    \item Rock Paper Scissors
    \item Running On Four
    \item Sack Race
    \item Sand Scuplting
    \item Segway
    \item Shoot Dance
    \item Shooting Handgun
    \item Shooting Shotgun
    \item Shuffle Dance
    \item Sling
    \item Slingshot
    \item Snow Angel
    \item Speed Stack
    \item Spinning Basketball
    \item Spinning Book
    \item Spinning Plate
    \item Stone Skipping
    \item Sword Swallowing
    \item Taichi Fan
    \item Taking Photo Camera
    \item Taking Selfie
    \item Tap Dancing
    \item Three Legged Race
    \item Throw Boomerang
    \item Throw Paper-Plane
    \item Tight-Rope Walking
    \item Trampoline
    \item Tug Of War
    \item Underarm Turn
    \item Walking On Stilts
    \item Whistle One Hand
    \item Whistle Two Hands
\end{enumerate}

\section{Composite Classes}
We list how \textit{Musical Instrument} and \textit{Sports/Athletics} classes form to become composite actions. We list indices of the classes for each composite action.

\subsection{Sports/Athletics}
\begin{enumerate}[itemsep=-2mm]
  \setcounter{enumi}{0}
    \item 49
    \item 79, 80
    \item 99
    \item 65, 66, 67
    \item 2
    \item 178, 179, 180, 181, 182
    \item 120, 121
    \item 39, 40, 41, 42
    \item 114
    \item 140
    \item 111, 112
    \item 25, 26, 27, 28, 29, 30, 31
    \item 60
    \item 56, 57, 58
    \item 156
    \item 144
    \item 59
    \item 150, 151, 152
    \item 168
    \item 167
    \item 102, 103
    \item 145
    \item 81, 82, 83
    \item 92
    \item 128, 129
    \item 50, 51
    \item 53, 54
    \item 138, 139
    \item 43
    \item 174, 175, 176, 177
    \item 183, 184, 185
    \item 201
    \item 8, 9, 10, 11, 12
    \item 142
    \item 149
    \item 1
    \item 32
    \item 62, 63, 64
    \item 141
    \item 109
    \item 104
    \item 122
    \item 110
    \item 38
    \item 100
    \item 157
    \item 37
    \item 197, 198, 199, 200
    \item 116
    \item 153, 154, 155
    \item 84
    \item 131
    \item 127
    \item 18, 19, 20, 21, 22, 23, 24
    \item 117, 118, 119
    \item 186, 187
    \item 160
    \item 169, 170, 171
    \item 158, 159
    \item 206, 207
    \item 13
    \item 172
    \item 133, 134, 135, 136, 137
    \item 123
    \item 124
    \item 205
    \item 5, 6, 7
    \item 86
    \item 208, 209, 210, 211, 212
    \item 113
    \item 202, 203, 204
    \item 166
    \item 105, 106, 107, 108
    \item 192, 193, 194, 195, 196
    \item 125, 126
    \item 61
    \item 173
    \item 143
    \item 85
    \item 188, 189
    \item 130
    \item 101
    \item 55
    \item 68, 69, 70, 71, 72, 73, 74, 75, 76, 77, 78
    \item 52
    \item 115
    \item 91
    \item 146, 147, 148
    \item 87, 88
    \item 44, 45
    \item 89, 90
    \item 3
    \item 190, 191
    \item 4
    \item 35, 36
    \item 34
    \item 33
    \item 14, 15, 16, 17, 46, 47, 48
    \item 93, 94, 95, 96, 97, 98
    \item 132
    \item 161, 162, 163, 164, 165
\end{enumerate}

\subsection{Musical Instruments}
\begin{enumerate}[itemsep=-2mm]
  \setcounter{enumi}{0}
    \item 369, 377, 379, 388, 390, 394, 395, 397, 398, 401, 402, 403, 406, 412, 413,399
    \item 371, 373, 376, 381, 382, 385, 387, 391, 396, 400, 404, 409, 414, 415, 416
    \item 370, 372, 374, 375, 378, 380, 383, 386, 389, 392, 405, 407, 408, 410, 411, 417, 418
    \item 368, 384, 393
\end{enumerate}

\section{HAA-COCO}

Here we list the classes in HAA-COCO. 

\begin{itemize}
    \item 1, 4, 5, 6, 7, 8, 9, 10, 11, 13, 14, 15, 16, 17, 18, 19, 20, 21, 22, 23, 32, 33, 34, 35, 36, 37, 45, 46, 47, 58, 60, 80, 81, 82, 86, 87, 88, 89, 99, 108, 110, 111, 115, 124, 125, 127, 128, 132, 133, 134, 135, 136, 139, 142, 160, 161, 162, 163, 164, 165, 182, 183, 184, 196, 197, 198, 199, 201, 202, 203, 212, 214, 235, 236, 237, 245, 246, 248, 249, 250, 251, 252, 263, 264, 265, 266, 267, 268, 276, 277, 278, 289, 298, 299, 305, 307, 311, 312, 313, 314, 316, 317, 318, 321, 325, 328, 330, 336, 337, 339, 345, 357, 358, 359, 360, 361, 367, 375, 376, 379, 381, 383, 384, 386, 388, 398, 400, 409, 410, 411, 412, 413, 414, 415, 427, 431, 434, 435, 443, 454, 480, 481, 487, 488
\end{itemize}

\section{Sample Videos} 
Figure~\ref{fig:supp_0} shows the first frame of a video in different classes.  Figure~\ref{fig:supp_2} lists diverse videos per class.

\section{Hierarchy}
Figure~\ref{fig:supp_1} shows the hierarchy of action classes in \textit{Sports/Athletics} area where the actions are grouped together with other actions in the same sports category.

{\small
\bibliographystyle{ieee_fullname}
\bibliography{egbib}
}